\pdfoutput=1
\documentclass[11pt]{article}

\usepackage{EACL2023}

\usepackage{times}
\usepackage{hyperref}
\usepackage{latexsym}
\usepackage{multirow}
\usepackage{graphicx}
\usepackage{amssymb}% http://ctan.org/pkg/amssymb
\usepackage{pifont}% http://ctan.org/pkg/pifont
\usepackage{caption}
\usepackage{subcaption}
\usepackage{stfloats}
\usepackage{bm}
\usepackage{fancyhdr}
\usepackage{arydshln}

\usepackage[T1]{fontenc}
\usepackage[utf8]{inputenc}

\usepackage{microtype}

\usepackage{inconsolata}

\usepackage[capitalize]{cleveref}
\crefname{section}{Sec.}{Secs.}
\Crefname{section}{Section}{Sections}
\Crefname{table}{Table}{Tables}
\crefname{table}{Tab.}{Tabs.}

\title{Multilingual Content Moderation: A Case Study on Reddit}
\author{
Meng Ye\textsuperscript{1}, Karan Sikka\textsuperscript{1}, Katherine Atwell\textsuperscript{2}, Sabit Hassan\textsuperscript{2}\\
{\bf Ajay Divakaran\textsuperscript{1}}, {\bf Malihe Alikhani\textsuperscript{2}} \\
  \textsuperscript{1}Center for Vision Technologies, SRI International \\
  \textsuperscript{2}Department of Computer Science, University of Pittsburgh \\
  \texttt{\{first.last\}@sri.com}, \texttt{\{kaa139, sah259, malihe\}@pitt.edu} \\
  }

\begin{document}
\maketitle
% DISTRIBUTION STATEMENT (requested by DARPA)
% \fancypagestyle{firststyle}
% {
%   \fancyhf{}
%   %\fancyfoot[C]{\footnotesize Page \thepage\ of \pageref{LastPage}}
%     \fancyfoot[C]{DISTRIBUTION STATEMENT A. Approved for public release; distribution is unlimited.}
%   \renewcommand{\headrulewidth}{0pt} % removes horizontal header line
% }
% \thispagestyle{firststyle}
% \pagestyle{firststyle}

\setlength{\abovecaptionskip}{6pt}
\setlength{\belowcaptionskip}{-3pt}
\begin{abstract}
Content moderation is the process of flagging content based on pre-defined platform rules.
% While prior works have put effort into building hate speech and offensive language datasets, they are not sufficient for solving content moderation. 
% Moderation requires detecting violation of community guidelines, which subsumes detection of offensive or hate speech, and changes dynamically across communities. 
% In this work we present a multilingual dataset of Reddit user comments collected from $56$ subreddits to study content moderation. 
% Our dataset contains rules for each subreddit and moderation annotation indicating whether a comment was flagged by the moderators for $\sim1.8$ million comments. 
% Along with the dataset we also provide insights about the data, underlying challenges and discuss the limitations of prior works in solving content moderation.
% We also provide experimental results of baseline NLP models and study the possibility of transferring models trained on English to non-English data. We believe that this dataset will help the community in building better AI moderators.
There has been a growing need for AI moderators to safeguard users as well as protect the mental health of human moderators from traumatic content.
While prior works have focused on identifying hateful/offensive language, they are not adequate for meeting the challenges of content moderation since
1) moderation decisions are based on violation of rules, which subsumes detection of offensive speech, and 2) such rules often differ across communities which entails an adaptive solution.
% ksikka- I shortend it. Point2 is also common with offensive speech.
% 1) moderation decisions are based on violation of rules, which subsumes detection of offensive speech,
% 2) it requires contextualized understanding of the conversation to decide whether a comment is offensive, on topic, or violating rules,
% and 3) it is affected by the biases of each individual moderator.
We propose to study the challenges of content moderation by introducing a multilingual dataset of \textbf{1.8 Million} Reddit comments spanning \textbf{56} subreddits in English, German, Spanish and French\footnote{\url{https://github.com/mye1225/multilingual_content_mod.git}}.
We perform extensive experimental analysis to highlight the underlying challenges and suggest related research problems such as cross-lingual transfer, learning under label noise (human biases), transfer of moderation models, and predicting the violated rule.
% gain insights into the the challenges and 
% provide the moderation labels and subreddit metadata that includes description, and community rules.
% Our dataset can not only be used to study content moderation, but also cross-lingual transfer and moderator biases.
Our dataset and analysis can help better prepare for the challenges and opportunities of auto moderation.
%\blfootnote{a footnote}
\end{abstract}
    
%%%%%%%%%%%%%%%%%%%%%%%%%%%%%%%%%%%%%%%%%%%%%%%%%%%%
%% Introduction
%%%%%%%%%%%%%%%%%%%%%%%%%%%%%%%%%%%%%%%%%%%%%%%%%%%%
\section{Introduction}
\begin{figure}[ht]
    \centering
    \includegraphics[width=0.99\linewidth]{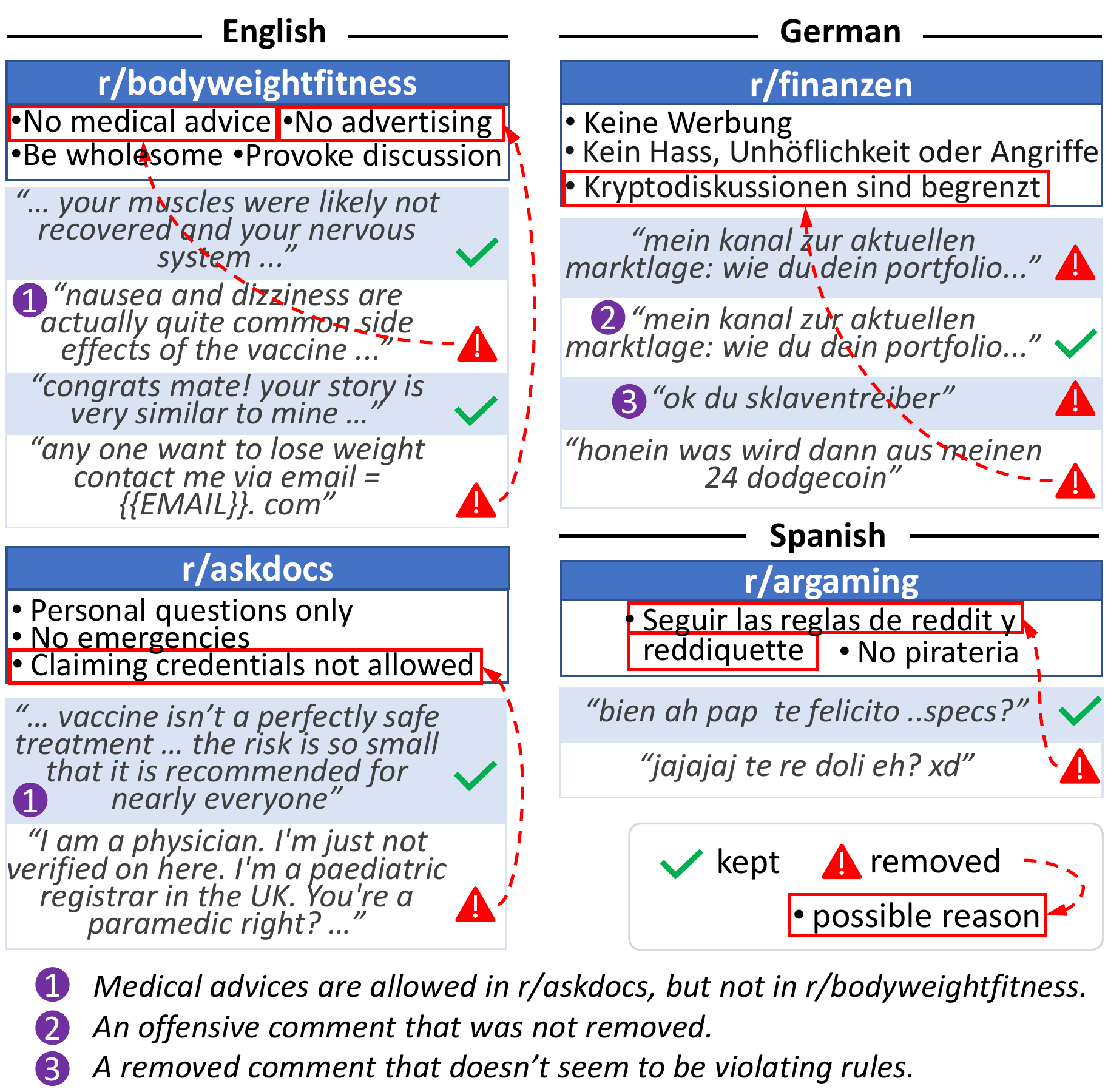}
    \caption{Content moderation on Reddit is challenging as it 1) depends on rules of each community, 2) requires contextualized understanding of the comments, and 3) is affected by moderator biases.}
    \label{fig:figure1}
\end{figure}
Being able to moderate user-generated content is critical for online social media platforms. 
Several platforms employ human moderators to monitor user content to prevent the spread of misinformation, adverse effects of hateful speech, fraud, etc ~\cite{geiger2010work, dosono2019moderation, wang2022governing, jhaver2017designing}. 
The moderators' task is to remove improper content and/or suspend users posting such content.
However, reviewing and moderating each user comment is practically infeasible due to limited resources, especially during time-critical and large-scale events. More importantly, such moderation work can cause damage to moderators' mental health due to burnout from volunteer work and exposure to harmful content %, navigating hierarchical structures, balancing unfulfilled expectations 
%among other reasons
\cite{gillespie2020content, roberts2014behind, dosono2019moderation, wohn2019volunteer}.

In recent years, researchers have put effort into collecting hate speech or offensive language datasets from social platforms such as Twitter~\cite{zampieri-etal-2019-predicting,gautam2020metooma,golbeck2017large}, YouTube~\cite{dinakar2011modeling}, or a mixture of those platforms~\cite{kennedy2017technology,bhattacharya2020developing}.
These datasets contain annotations of hateful speech and are used to train NLP systems to remove harmful comments~\cite{pamungkas2020you, chung-etal-2019-conan, ranasinghe2021multilingual}. We refer to this problem as Offensive Language Identification (OLI).
Instead of simply removing such content or suspending users, some works propose countering hate speech with discourse-aware style transfer~\cite{Atwell2022APPDIAAD}, expert-based counter-narratives~\cite{chung-etal-2019-conan}, by learning to intervene with conversational context~\cite{qian2019benchmark}, or by proposing a holistic conceptual framework~\cite{chaudhary2021countering}.

However, despite remarkable progress in OLI, it is not sufficient to tackle content moderation for two key reasons: 
(1) OLI is a subset of content moderation, as the latter involves flagging content that is not only hateful but also violates the rules of the platform. For example, 
%subreddit ``r/stock'' does not allow comments that are low-effort, off-topic or self-promotional in addition to Reddit sitewise rules\footnote{\url{https://www.redditinc.com/policies/content-policy}}.
Reddit has site-wise rules that not only requires users to be civil, but also prohibits actions such as posting illegal content, spamming, and revealing personal information\footnote{\url{https://www.redditinc.com/policies/content-policy}}.
(2) Prohibiting offensive/hateful speech is universal to all communities~\cite{chandrasekharan2018internet}, while content moderation requires a system to be adaptive to rules that change dynamically across different communities.
It is often the case that content that is allowed in one subreddit might be disallowed in another.
For example, it is allowed to post medical advice on {\tt r/askdocs} while it is prohibited on {\tt r/bodyweightfitness}.
%Since Reddit is an example of a platform that relies on individual community moderators, unaffiliated with Reddit itself, to set and enforce community guidelines.
We thus believe it is useful to study Reddit comments from a moderation perspective as we can collect data from multiple communities and measure the effectiveness of such systems in adapting to community guidelines.

In this work, we aim to bridge this gap by introducing a dataset of Reddit comments collected from $56$ communities for studying content moderation. We propose a challenge for content moderation and demonstrate the limitations of the state-of-the-art models. Our dataset will be available to researchers who agree to the terms and conditions of our data-sharing policy approved by the ethical committee of our institution.
% The dataset consists of $1.8$ million comments together with moderation labels, that specify if the comment was kept or removed by the moderator. We also provide the meta-data for each subreddit including its name, description, and rules. Apart from English, we also collected comments from subreddits that use German, French, and Spanish as their first languages.
% We first describe the procedure for selecting these subreddits and then collecting data from them.
% We then provide analysis on the offensiveness distribution of the collected data and show that OLI models are not suitable for moderation because they fail to capture content that were removed but are non-offensive. We also present baseline performance of different NLP models trained on the collected moderation data under different cross-lingual settings. Finally we provide our insights on what makes content moderation a challenging task and potential opportunities with the proposed dataset. \textbf{Need to list contributions in a better manner and also other things like ethical etc.}
Our contributions are:
\begin{itemize}
    \item We propose a multilingual dataset for content moderation. The dataset consists of \textbf{1.8 million} comments from \textbf{56} subreddits with moderation labels, that specify whether a comment was kept or removed by the moderator. We also provide the meta-data for each subreddit that includes its name, description, and rules.
    \item We show that existing offensive speech datasets are not suitable for the content moderation task because only a small portion of the removed comments are offensive. As such, the models trained on OLI datasets fail to identify removed comments that are non-offensive.
    \item We study the performance of models trained on moderation data under different cross-lingual settings including multilingual-language model, translate-train, and translate-test.
    \item We provide insights on what makes content moderation a challenging task and discuss potential research problems that can be explored with our proposed dataset.
\end{itemize}

%%%%%%%%%%%%%%%%%%%%%%%%%%%%%%%%%%%%%%%%%%%%%%%%%%%%
%% Related works
%%%%%%%%%%%%%%%%%%%%%%%%%%%%%%%%%%%%%%%%%%%%%%%%%%%%
\section{Related Work}

\begin{table}
\centering
\small
% \begin{tabular}{c|c|c|c|c}
%     \hline
%     Dataset & Platform & Type & Size & Language \\
%     \hline
%     OLID\cite{zampieri-etal-2019-predicting} & Twitter & OFF & $14$K & English \\
%     SWAD\cite{pamungkas2020you} & Twitter & SW & $1.5$K & English \\
%     CONAN\cite{chung-etal-2019-conan} & Facebook/Synthetic & HS & $4$K & Multilingual \\
%     OffensEval-2020\cite{zampieri-etal-2020-semeval} & Twitter & OFF & $9$M & Multilingual \\
%     Macro norms\cite{chandrasekharan2018internet} & Reddit & MOD & $4.6$M & English \\
%     \hline
%     Ours & Reddit & MOD & $1.8$M & Multilingual \\
%     \hline
% \end{tabular}
\begin{tabular}{@{}lllrr@{}}
    \hline
    {\bf Dataset} & {\bf Platform} & {\bf Type} & {\bf Size} & {\bf Language} \\
    \hline
    OLID & T & O\&H & $14$K & en \\
    SWAD & T & SW & $1.5$K & en \\
    OffensEval & T,R,F & O\&H & $9$M & en, da, tr, ar, el \\
    HatEval & T & O\&H & 19.6K & en, es \\
    CDUC & Y & CYB & 4.6K & en \\
    CONAN & T & O\&H & $15$K & en, fr, it \\
    Norms & R & MOD & $4$M & en \\
    \hline
    Ours & R & MOD & $1.8$M & en, de, es, fr \\
    \hline
\end{tabular}
\caption{Comparison with published datasets. OLID~\cite{zampieri-etal-2019-predicting}. SWAD~\cite{pamungkas2020you}. OffensEval-2020~\cite{zampieri-etal-2020-semeval}. HatEval~\cite{basile2019semeval}. CONAN~\cite{chung-etal-2019-conan}. CDUC~\cite{dadvar2013improving}. Norms~\cite{chandrasekharan2018internet}. Platforms: (T)witter, (F)acebook, (R)eddit, (Y)outube. Types: (O)ffensive and (H)ate speech, (SW)earing, (CYB)er Bullying, (MOD)eration. }
\label{tab:related_datasets}
\end{table}

%%%%%%%%%%%%%%%%%%%%%%%%%%%%%%%%%%%%%%%%%%%%%%%%%%%%
%% Related works - Offensive Language and Hate Speech datasets
%%%%%%%%%%%%%%%%%%%%%%%%%%%%%%%%%%%%%%%%%%%%%%%%%%%%
\subsection{Offensive Language and Hate Speech Datasets}
Many existing works that have collected user comments from online social platforms. \citet{zampieri-etal-2019-predicting} proposed Offensive Language Identification Dataset (OLID), which models not only different types of offensive language, but also the target of offensive messages in a hierarchical structure. The Swear Words Abusiveness Dataset (SWAD) developed by~\citet{pamungkas2020you} is a collection of tweets selected from OLID that focuses on predicting abusiveness of a swear word in a tweet context.
\citet{chung-etal-2019-conan} created a multilingual dataset with hate speech/counter-narrative pairs provided by experts. The idea is to fight online hate speech content with informed textual responses instead of the standard method of removing content or suspending users.
OffensEval~\cite{zampieri-etal-2020-semeval} is an offensive language identification challenge that has attracted multiple research teams.
It is based on the same hierarchical three-level annotations from the aforementioned OLID dataset in multiple languages. CAD~\cite{vidgen-etal-2021-introducing} is a recent recently proposed dataset of Reddit posts and comments with manually annotated two-level abusive categories.
A brief comparison between these datasets and ours is shown in \autoref{tab:related_datasets}.

All the mentioned works focus on detecting offensive speech.
However, online content moderation involves detection of comments that violate community rules in addition to those that are offensive.
For example, moderators will often remove comments that are self-promoting, spamming, or off-topic because they do not provide useful information and are harmful for the communication environment (e.g., \textit{``I can help you, see my youtube channel''}). 
% ksikka- removed next line because it has the same content as the previous line
%Many subreddits also have there own specific rules and any comments violating those will be removed. For example, ``r/stocks'' does not allow discussions of cryptocurrencies unrelated to stocks so a comment like ``Bitcoin is bullish now buy it'' will be removed.
In both of these cases, a model trained to detect hate speech will likely fail.

%%%%%%%%%%%%%%%%%%%%%%%%%%%%%%%%%%%%%%%%%%%%%%%%%%%%
%% Related works - Content Moderation datasets
%%%%%%%%%%%%%%%%%%%%%%%%%%%%%%%%%%%%%%%%%%%%%%%%%%%%
\subsection{Reddit Rules and Content Moderation}
A pre-existing related work by \citet{chandrasekharan2018internet} trained 100 subreddit classifiers on removed/un-removed comments, and used clustering analysis to discover three types of \emph{implicit norms} from the removed comments: macro norms that are universal to all subreddits (e.g., hate speech, personal attacks), meso norms that applies to subgroups (e.g., meme responses), and micro norms that are specific to individual subreddits (e.g., offering commerce tips). Our study is different in that we focus on moderation task that depends on \emph{explicit community rules} written by moderators.
Another work~\cite{samory2021positive} studied the problem of identifying comments \emph{approved} by moderators. They found that approved comments and removed ones actually share many traits such as toxicity and insults, and that it is hard to distinguish them.
\citet{fiesler2018reddit} studied rules from $1,000$ subreddits and found that rules are highly dependent on the context of each individual subreddit while sharing common characteristics across the platform.

One key limitation of these studies is that they only focused on subreddits in the English language and discarded all non-English subreddits. This could be because most data on Reddit is in English.
In our work we also select subreddits in non-English languages, i.e. German, French, and Spanish. We then study the possibility of transferring a moderation model trained on subreddits in English (which generally has more content). We argue that a good AI moderator should not only focus on data in English language, but also leverage that knowledge to improve performance on other low-resource languages.
While \citet{hassan2022studying} study \textit{human} moderation bias across different languages and cultures, our work focuses on the problem of \textit{automated} moderation.

%%%%%%%%%%%%%%%%%%%%%%%%%%%%%%%%%%%%%%%%%%%%%%%%%%%%
%% Dataset
%%%%%%%%%%%%%%%%%%%%%%%%%%%%%%%%%%%%%%%%%%%%%%%%%%%%
\section{Dataset}
\label{sec:dataset}

%We now introduce our dataset and describe the process of selecting the subreddits, the data collection pipeline and also provide statistics for the dataset.

\subsection{Subreddit selection}
We collected data from $56$ subreddits\footnote{A full list of all 56 subreddits can be found in \autoref{tab:subreddit_list}.} based on the following criteria:
  
\noindent \textbf{Wide range of topics}: Generally the popularity (and thus data points) of a subreddit depends on its topic. Being able to cover many topics will enable us to better estimate the generalizability of machine learning models on this task. The topics include news, politics, finance, sports, electronics, etc.
    
\noindent \textbf{Subreddits on similar topics}: We chose subreddits with similar topics to enable inquiry into questions relating to transferability of moderation models. For example, ``Do models trained on one subreddit transfer to other subreddits?'', ``will models transfer better between subreddits on similar topics?'', and ``which subreddit should I train the model on so that it could also be better adapted to X?''. Some example of similar subreddits are {\tt r/news} and {\tt r/worldnews}, {\tt r/finance} and {\tt r/personalfinance}, {\tt r/anime}                                  and {\tt r/naruto}, {\tt r/games} and {\tt r/xboxseriesx}.
    %The reason why we choose these paired subreddits is that it will allow us to study the transferrability between them and answer questions like ``Do models trained on one subreddit transfer to other subreddits?'', ``will models transfer better between subreddits on similar topics?'', and ``which subreddit should I train the model on so that it could also be better adapted to X?''.

\noindent \textbf{Multilingual data}: Most groups on Reddit are in English, but there are also some in other languages such as Spanish, German, French, Arabic. To extend moderation models to multilingual settings we also selected a number of non-English subreddits. For example French ({\tt r/quebec}, {\tt r/france}, {\tt r/moi\_dlvv}), German ({\tt r/de}, {\tt r/finanzen}, {\tt r/ich\_iel}), and Spanish ({\tt r/argentina}, {\tt r/gaming}).

%%%%%%%%%%%%%%%%%%%%%%%%%%%%%%%%%%%%%%%%%%%%%%%%%%%%
%% Dataset - Data collection pipeline
%%%%%%%%%%%%%%%%%%%%%%%%%%%%%%%%%%%%%%%%%%%%%%%%%%%%
\subsection{Data Collection Pipeline}
\label{data_pipeline}
We built our data collection pipeline based on the Python Reddit API Wrapper (PRAW)
\footnote{\url{https://github.com/praw-dev/praw}}, which streams
comments and submissions in real time from multiple subreddits. 
For each data record we store important fields such as the ID, author, posting time, and comment body. 
We follow a two step approach where we first scrape data for a week, and then check if the comment/submission has been removed or retained by the moderator or if it has been deleted by the author. We use this two-step approach since once the comment is removed/deleted it does not retain its original content.
% Simplified since we can provide more details
%We keep this scraping process running for around a week, after which we start another process to check the status of all the comments/submissions that have been collected during the past week.
% The checker process retrieves each data record from Reddit based on its id, and then check the retrieved comment/submission text body if it is `[removed]' (which means the original post has been removed by the moderator), `[deleted]' (deleted by the author), or otherwise (might be edited or not).
\autoref{fig:pipeline} in the appendix shows the overall procedure of our data collection pipeline.

%%%%%%%%%%%%%%%%%%%%%%%%%%%%%%%%%%%%%%%%%%%%%%%%%%%%
%% Dataset - Data statistics
%%%%%%%%%%%%%%%%%%%%%%%%%%%%%%%%%%%%%%%%%%%%%%%%%%%%
\subsection{Dataset Overview}
We use the data collection pipeline to collect $\sim 1.8$ Million comments in a time span of three weeks.  

For benchmarking, we split all of the English data randomly into $90\%/5\%/5\%$ as train, validation, and test subsets. 
We chose all the non-English data to be part of the test set to study the cross-lingual transferability of models trained on data in English.
\autoref{tab:statistics} lists the number of subreddits, total number of comments, and the percentage of removed comments for all the data splits.

\begin{table}[t]
\small
    \centering
    \begin{tabular}{@{}llrrr@{}}
        \hline
        {\bf split} & {\bf language} & {\bf \#sub} & {\bf \#comments} & {\bf removal} \\
        \hline
        \multicolumn{5}{c}{Training data} \\
        %\hline
        en-train & English & $48$ & $1,347,611$ & $1.87\%$ \\
        en-val & English & $48$ & $74,885$ & $1.83\%$  \\
        \hline
        \multicolumn{5}{c}{Test data} \\
        %\hline
        de & German & $3$ & $177,046$ & $0.86\%$ \\
        es & Spanish & $2$ & $95,586$ & $0.66\%$ \\
        en & English & $48$ & $74,893$ & $1.85\%$ \\
        fr & French & $3$ & $49,780$ & $0.23\%$ \\
        \hline
    \end{tabular}
    \caption{Number of subreddits, comments, and percentage of removed comments in each data split.}
    \label{tab:statistics}
\end{table}

English data makes up  a large proportion of the entire dataset, and is collected from $48$ subreddits on different topics.
For non-English languages, there are significantly fewer subreddits and we choose the most popular ones for each language.
This dataset is highly imbalanced in that only a small proportion of the comments are removed by the moderators: English data has a removal rate of around $1.8\%$ and for German, Spanish, and French it is $0.86\%$, $0.66\%$, and $0.23\%$, respectively.
This makes the task of identifying moderated comments very challenging. For comparison, the proportion of offensive examples is around $33\%$ in OLID, $12.5\%\sim28.4\%$ across different languages in OffensEval-2020~\cite{zampieri-etal-2020-semeval},  and $28.4\%\sim50.0\%$ in HASOC-2020~\cite{mandl2020overview}.
We argue that the high percentage of offensive examples in those datasets makes them less suitable for real world applications because these offensive posts make up only a small share of moderated content.

We also observe that different subreddits have different rates of removal, from the lowest of $0.0\%$ to the highest of $21.74\%$.
These differences are likely explained by the discrepancies in topic and subreddit rules, as well as the level to which the moderators enforce these rules.
However, the overall rate of removal is quite low ($1.84\%$) compared to existing datasets mentioned above. We show the statistics of the top 5 active subreddits in \autoref{fig:subreddits-en-top5} (refer to \autoref{fig:subreddits-en} in the appendix for more details).

\begin{figure}[t]
    \centering
    \includegraphics[width=\linewidth]{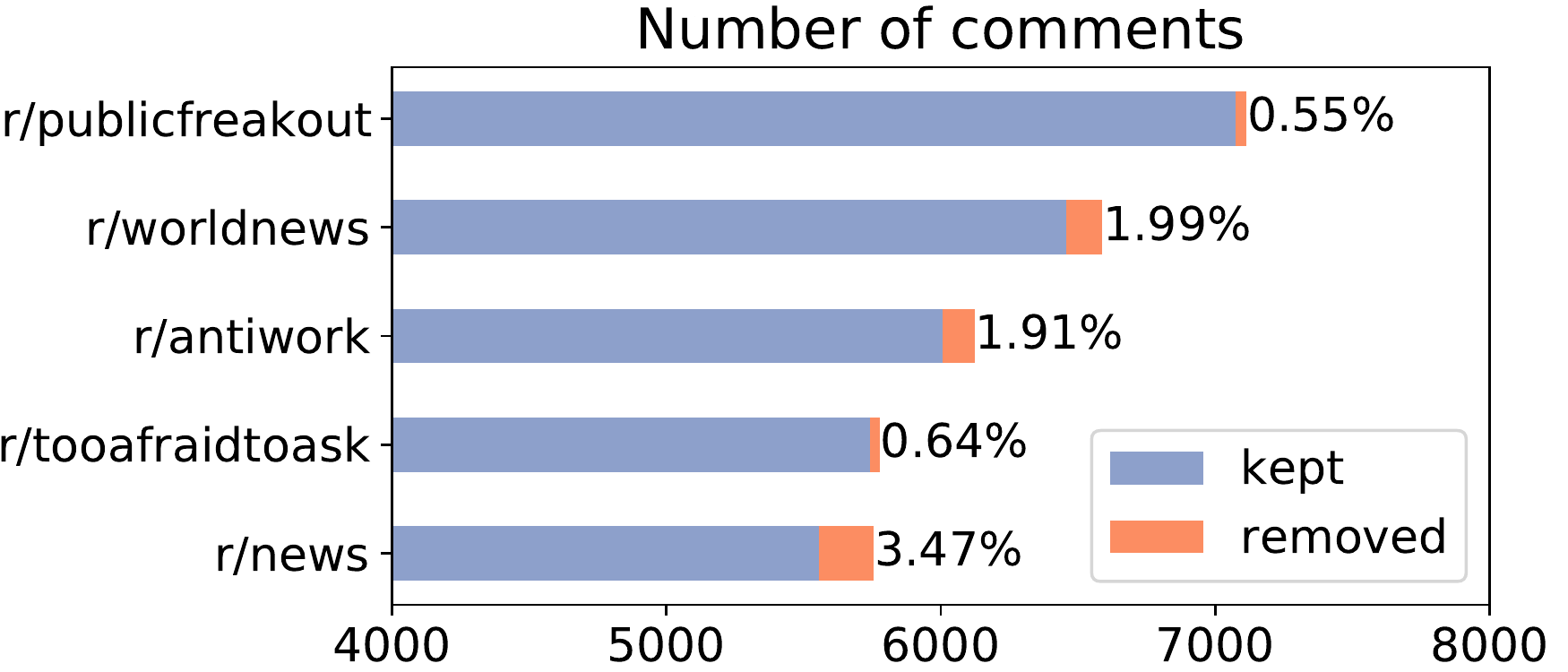}
    \caption{Number of kept and removed comments in the top 5 subreddits (\% removed comments in also shown).}
    \label{fig:subreddits-en-top5}
\end{figure}

% \begin{figure}[t]
%     \centering
%     \begin{subfigure}[b]{\linewidth}
%         \centering
%         \includegraphics[width=\textwidth]{figures/alldata-en-val-en.pdf}
%         \caption{English}
%         \label{fig:subreddits-en}
%     \end{subfigure}
    
%     \begin{subfigure}[b]{0.3\linewidth}
%         \centering
%         \includegraphics[width=\linewidth]{figures/alldata-de-de.pdf}
%         \caption{German}
%         \label{fig:subreddits-de}
%     \end{subfigure}
%     \begin{subfigure}[b]{0.3\linewidth}
%         \centering
%         \includegraphics[width=\linewidth]{figures/alldata-es-es.pdf}
%         \caption{Spanish}
%         \label{fig:subreddits-es}
%     \end{subfigure}
%     \begin{subfigure}[b]{0.3\linewidth}
%         \centering
%         \includegraphics[width=\linewidth]{figures/alldata-fr-fr.pdf}
%         \caption{French}
%         \label{fig:subreddits-fr}
%     \end{subfigure}
%     \caption{Distribution of the number of comments in each subreddit for each language.}
%     \label{fig:subreddits}
% \end{figure}

Each subreddit is associated with a set of rules that are enforced by the moderators, which are normally displayed in the \emph{About} section. Typically, each rule has a short title and a longer description to explain in more detail the types of content that are prohibited. For each subreddit, we have collected all of its rules including titles and descriptions (see \autoref{fig:subreddits-rules} in the Appendix for the distribution of the number of rules in our dataset).

%%%%%%%%%%%%%%%%%%%%%%%%%%%%%%%%%%%%%%%%%%%%%%%%%%%%
%% Dataset - Manual Analysis of Offensiveness
%%%%%%%%%%%%%%%%%%%%%%%%%%%%%%%%%%%%%%%%%%%%%%%%%%%%

\subsection{Manual Analysis of Offensiveness}
\label{sec:manual_offensiveness}

\begin{figure}[t]
    \centering
    \includegraphics[width=0.49\linewidth]{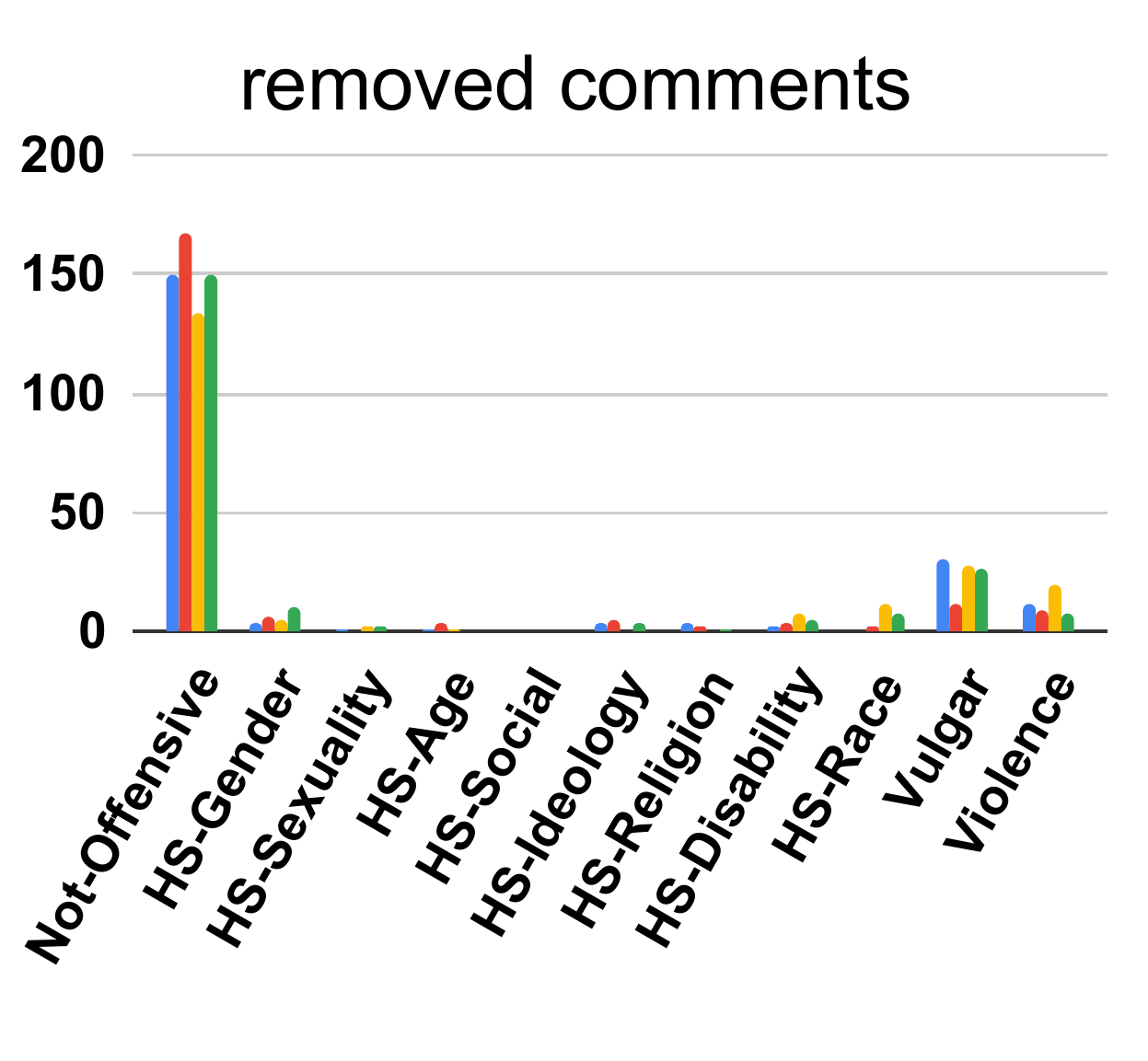}
    \includegraphics[width=0.49\linewidth]{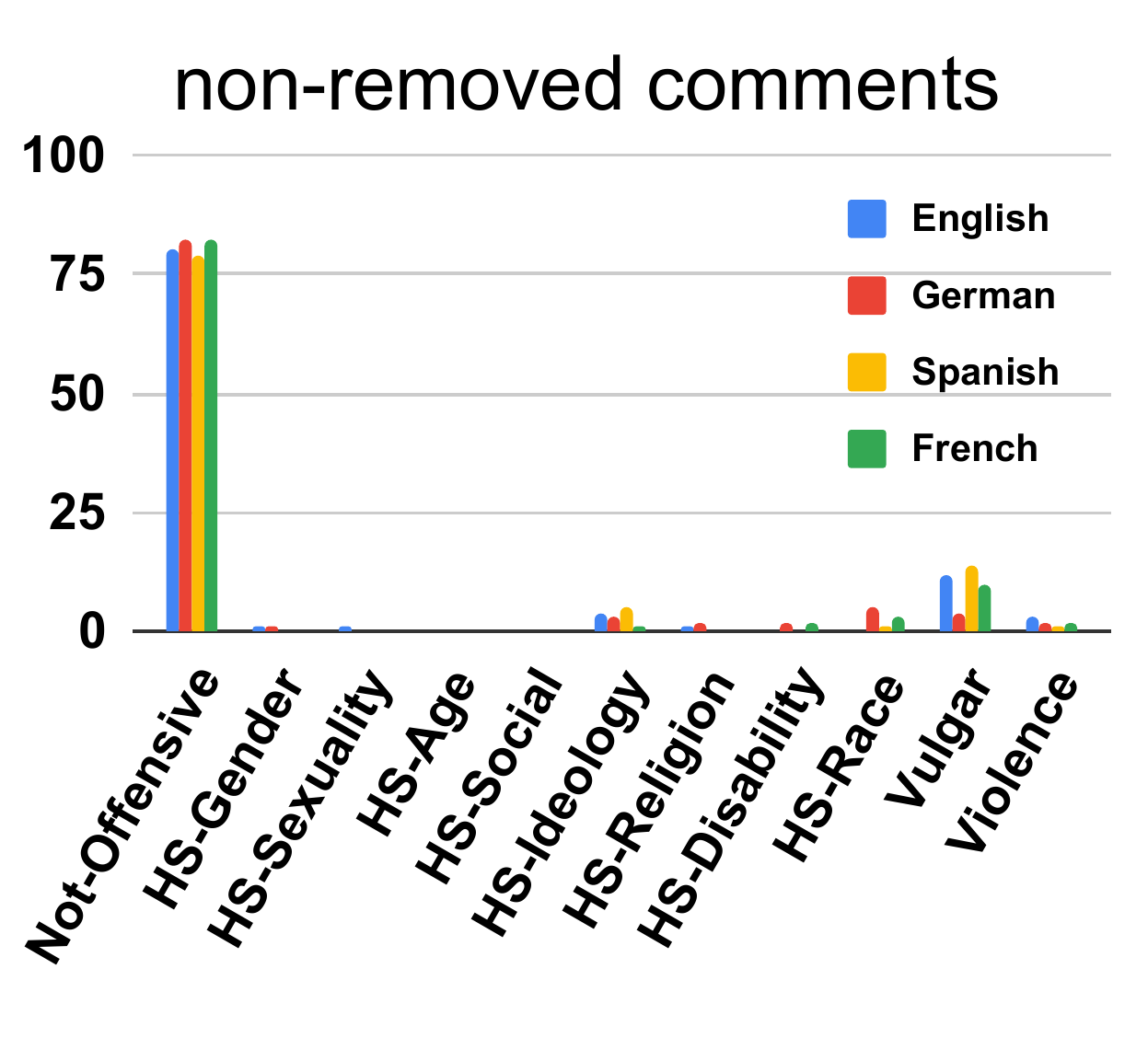}
    \caption{Distribution of different types of offensive speech for Reddit comments.}
    \label{distribution}
\end{figure}

To better understand the level of offensiveness in the content moderation task, we manually annotated $1,238$ comments (around $200$ removed and $100$ unremoved examples for each of the four languages) using the fine-grained taxonomy of offensiveness presented in \citet{Mubarak2022EmojisAA}, with the addition of the categories of sexuality and age. The distribution of comments for different categories is shown in \autoref{distribution}. We observed that most of the comments ($71.86\%$ of removed and $80.115\%$ of non-removed) are not offensive. Many of these non-offensive comments criticized the corresponding subreddit or disagreed with the views or goals of the subreddit. For instance, the comment \textit{"i was disappointed when i got the halo console..."} was removed from the {\tt r/halo} subreddit. The comment is not offensive but could have been removed due to its criticism of a product supported by the {\tt r/halo} subreddit. Since these rules, views and goals can vary significantly across subreddits, it explains why content moderation is hard for machine learning classifiers and our moderation classifiers achieve lower scores compared to offensiveness classifiers.
%We also observe that among different types of offensiveness in the removed comments, vulgar comments were most prominent.
We also see that the distributions of different types of offensive speech are similar across both removed and non-removed comments as well as the four different languages. 
To conclude, moderating online content is a very different task compared to identifying offensive language and hate speech and a new dataset is needed to train models and study automated content moderation.

%%%%%%%%%%%%%%%%%%%%%%%%%%%%%%%%%%%%%%%%%%%%%%%%%%%%
%% Experiment
%%%%%%%%%%%%%%%%%%%%%%%%%%%%%%%%%%%%%%%%%%%%%%%%%%%%

\section{Experiments}

% In this section we performed a more detailed analysis on the dataset.
% We start by training the moderation classifier on English data only.
% Since we are interested in transferability between languages,
% we evaluate the classifier on non-English comments.
% \emph{translation-train} and \emph{translate-test} settings are also added
% as these are common baselines for multilingual tasks.
% After getting a sense of how well supervised learning systems perform on the data,
% we examine the data qualitatively and show some examples that reveal some issues of this dataset,
% which reflect real-world challenges when collecting data and building auto moderation systems for social media platforms.

%%%%%%%%%%%%%%%%%%%%%%%%%%%%%%%%%%%%%%%%%%%%%%%%%%%%
%% Experiment - Task and Metrics
%%%%%%%%%%%%%%%%%%%%%%%%%%%%%%%%%%%%%%%%%%%%%%%%%%%%
%\subsection{Task and Metrics}
We study the task of content moderation on Reddit by formulating it as a binary classification task: given a user comment $x$ predict whether it should be removed by moderators using $y=f(x)$, where $f$ is the classifier, and $y$ is the removal probability. We set the binary labels based on whether the comment was removed or retained by the moderator (\autoref{data_pipeline}).
% We experiment with different types of classifiers $f$-- offensive language detection models and training $f$ using our Reddit dataset.
We report both AUC (area under the ROC curve) and F1 scores.

\subsection{Evaluation with Existing Offensive Language Identification Models}
\label{existing}
A naive solution to moderation is to cast this problem as offensive language detection and directly use classifiers trained on OLI datasets.
We thus evaluate models trained on OLI datasets for moderation.
This experiment will help in answering two questions: 1) can these models recognize offensive comments in the context of Reddit content moderation?, and 2) do these models miss comments that should be removed but are not-offensive? Being able to answer these question will help us understand whether content moderation can be solved by using OLI or it is a necessity to collect data tailored for moderation.
\begin{figure}[t]
    \centering
    \includegraphics[width=0.59\linewidth]{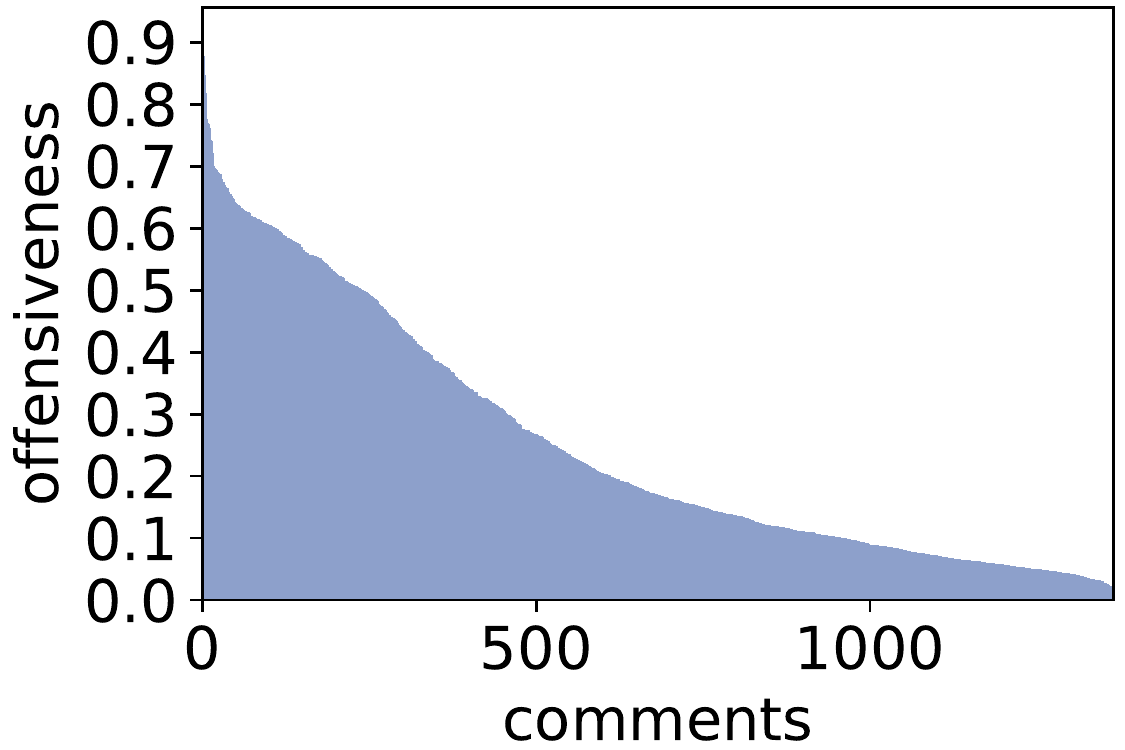}
    \includegraphics[width=0.39\linewidth]{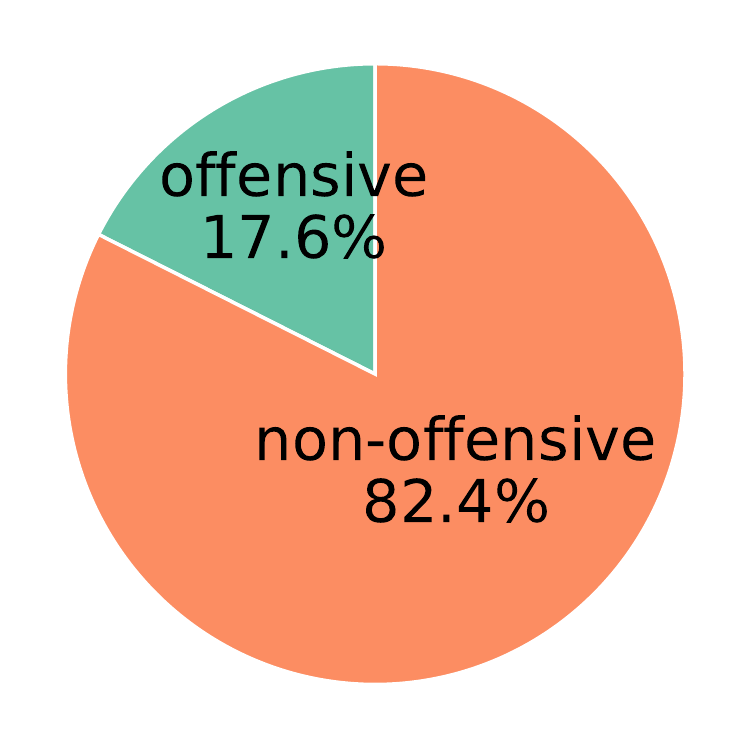}
    \caption{Distribution of offensiveness scores for removed comments (left plot shows sorted scores). Most of the removed comments are not very offensive. Only $18\%$ of them get offensive scores higher than $0.5$.}
    \label{fig:offensive_distribution}
\end{figure}
We investigate the above question using three models: RoBERTa (base)~\cite{liu2019roberta} finetuned on the HatEval~\cite{basile2019semeval} and OLID dataset~\cite{zampieri-etal-2020-semeval}, and an XLM-RoBERTa model fine-tuned on OLID.  
We compute the moderation score for each comment by averaging the output probabilities from these models.

\begin{table}[t]
    \centering
    \small
    \begin{tabular}{@{}llrrr@{}}
        \hline
        {\bf Model} & {\bf Tr data} & {\bf T-ori} & {\bf R-off} & {\bf R-mod} \\
        \hline
        RoBERTa & HatEval & $48.72$ & $44.12$ & $50.59$\\
        RoBERTa & OLID & $81.25$ & $79.20$ & $47.76$\\
        XLM-RoBERTa & OLID & $75.81$ & $75.82$ & $48.45$\\
        \hline
    \end{tabular}
    \caption{Offensive language/hate speech classifiers with their original training (Tr) Twitter dataset and macro-F1 scores on the test split of the same dataset (T-ori), a subset of our Reddit data with offensive labels (R-off), and our Reddit data with moderation labels (R-mod).}
    \label{tab:offensive_metric}
\end{table}
% For each example comment, we compute the offensiveness score for each comment (as a probabbility) from these models and use their average as the final moderation score.

\autoref{fig:offensive_distribution} shows the distribution of offensiveness scores of the removed comments. The left and the right plot show the sorted scores and the percentage of offensive and non-offensive comments using a threshold of $0.5$, respectively. We observe that most of the removed comments ($82.4\%$) are not offensive, and also note that this number matches with the manual analysis in \autoref{sec:manual_offensiveness}.

%We show the qualitative examples in \autoref{tab:offensiveness_examples}. 
By qualitatively studying these examples we observe that 1) OLI models can detect offensiveness in the comments, and 2) not all removed comments are offensive, and OLI models tend to miss these. For instance, a comment stating \textit{"try contacting drewcybersupport on instagram about ..."} from {\tt r/minecraft} was likely removed due to self-promotion (the average offensiveness of this comment is $0.01$). More examples can be found in \autoref{tab:offensiveness_examples}.
% , and the reason for "thanks for the good constructive conversation ..." from r/feminism being removed could be that it provides no other information except for expressing thanks and that /violates some meso norms~\cite{chandrasekharan2018internet}.
% These 
We also evaluate the performance of these models on three datasets: the original offensive datasets that they were fine-tuned on (T-ori), a subset of our collected Reddit data with manual offensiveness annotations (R-off), and our proposed moderation dataset (R-mod).
From the results in \autoref{tab:offensive_metric} we can observe that: 1) these models achieve good performance\footnote{The model fine-tuned on the HatEval dataset has a low F1 score possibly because it focused on hate speech targeted at immigrants and women instead of generic OLI.} on both Twitter and Reddit data when the task is to identify offensive language, and 2) their performance drops significantly when evaluated on the moderation task.
% summarizes each model and the dataset it was trained on as well as the the AUC it achieves on the English-test split of our moderation data.
%We observe that
% Although these classifiers are able to discriminate removed/non-removed comments better than chance, the best performing model only achieves an AUC of $61.1$.

Based on the results from both our qualitative and quantitative analyses, we conclude that there is a significant mismatch between the goals of content moderation and offensive language identification. We thus believe that it is necessary to have a moderation-specific dataset to train classifiers for content moderation.

\begin{table*}[t]
\setlength\tabcolsep{4pt}
    \centering
    \small
    \begin{tabular}{@{}lllcccccccccc@{}}
         \hline
         \multirow{2}{*}{{\bf Setting}} & \multirow{2}{*}{{\bf Language}} & \multirow{2}{*}{{\bf Encoder}} && \multicolumn{4}{c}{{\bf AUC}} && \multicolumn{4}{c}{{\bf F1-macro}}\\
         \cline{5-8} \cline{10-13} 
         & & && en & de & fr & es && en & de & fr & es \\
         \hline
         MLLM & multilingual & XLM-RoBERTa && $66.83$ & $69.42$ & $64.25$ & $63.90$ && $49.60$ & $49.78$ & $49.94$ & $49.83$ \\
         \hdashline[1pt/2pt]
         \multirow{3}{*}{Translate-train} & German & bert-base-german-uncased && $60.33$ & $63.34$ & - & - && $49.53$ & $49.78$ & - & - \\
         & French & flaubert-base-uncased && $53.47$ & - & $49.96$ & - && $49.53$ & - & $49.94$ & - \\
         & Spanish & bert-base-spanish-uncased && $64.16$ & - & - & $63.29$ && $49.58$ & - & - & $49.99$ \\
         \hdashline[1pt/2pt]
         Translate-test & English & roberta-base && $67.38$ & $71.23$ & $71.61$ & $64.33$ && $49.66$ & $50.01$ & $50.36$ & $50.22$ \\
         \hline
    \end{tabular}
    \caption{Experimental results on moderation classifier under three settings: 1) MLLM: multilingual language model embeddings, 2) translate-train, and 3) translate-test.}
    \label{tab:classification-original}
\end{table*}

% \begin{table*}[t]
% \setlength\tabcolsep{4pt}
%     \centering
%     \small
%     \begin{tabular}{@{}lllcccccccccc@{}}
%          \hline
%          \multirow{2}{*}{{\bf Setting}} & \multirow{2}{*}{{\bf Language}} & \multirow{2}{*}{{\bf Encoder}} && \multicolumn{4}{c}{{\bf AUC}} && \multicolumn{4}{c}{{\bf F1-macro}} \\
%          \cline{5-8} \cline{10-13}
%          & & && en & de & fr & es && en & de & fr & es \\
%          \hline
%          MLLM & multilingual & XLM-RoBERTa && $66.10$ & $69.93$ & $68.41$ & $59.74$ && $59.80$ & $64.47$ & $62.29$ & $56.80$ \\
%          \hdashline[1pt/2pt]
%          \multirow{3}{*}{Translate-train} & German & bert-base-german-uncased && $59.91$ & $64.34$ & - & - && $58.61$ & $62.25$ & - & - \\
%          & French & flaubert-base-uncased && $53.97$ & - & $54.12$ & - && $49.51$ & - & $54.64$ & - \\
%          & Spanish & bert-base-spanish-uncased && $62.97$ & - & - & $60.19$ && $58.48$ & - & - & $57.40$ \\
%          \hdashline[1pt/2pt]
%          Translate-test & English & roberta-base && $66.99$ & $70.67$ & $71.53$ & $62.04$ && $62.24$ & $64.75$ & $64.13$ & $58.17$ \\
%          \hline
%     \end{tabular}
%     \caption{Experimental results with balanced data splits from our moderation dataset.}
%     \label{tab:classification-balanced}
% \end{table*}

\begin{table*}[t]
\setlength\tabcolsep{4pt}
    \centering
    \small
    \begin{tabular}{@{}lllcccccccccc@{}}
         \hline
         \multirow{2}{*}{{\bf Setting}} & \multirow{2}{*}{{\bf Language}} & \multirow{2}{*}{{\bf Encoder}} && \multicolumn{4}{c}{{\bf AUC}} && \multicolumn{4}{c}{{\bf F1-macro}} \\
         \cline{5-8} \cline{10-13}
         & & && en & de & fr & es && en & de & fr & es \\
         \hline
         MLLM & multilingual & XLM-RoBERTa && $66.14$ & $70.10$ & $68.32$ & $59.81$ && $59.97$ & $64.46$ & $61.15$ & $56.78$ \\
         \hdashline[1pt/2pt]
         \multirow{3}{*}{Translate-train} & German & bert-base-german-uncased && $61.10$ & $65.90$ & - & - && $57.24$ & $61.15$ & - & - \\
         & French & flaubert-base-uncased && $53.64$ & - & $53.71$ & - && $51.99$ & - & $55.10$ & - \\
         & Spanish & bert-base-spanish-uncased && $63.08$ & - & - & $60.52$ && $59.33$ & - & - & $56.87$ \\
         \hdashline[1pt/2pt]
         Translate-test & English & roberta-base && $66.94$ & $70.66$ & $71.57$ & $61.96$ && $61.96$ & $64.16$ & $64.87$ & $58.11$ \\
         \hline
    \end{tabular}
    \caption{Experimental results with balanced data splits from our moderation dataset.}
    \label{tab:classification-balanced}
\end{table*}

%%%%%%%%%%%%%%%%%%%%%%%%%%%%%%%%%%%%%%%%%%%%%%%%%%%%
%% Experiment - Using moderation training data
%%%%%%%%%%%%%%%%%%%%%%%%%%%%%%%%%%%%%%%%%%%%%%%%%%%%

\subsection{Evaluation with Models Trained on Proposed Dataset}
We use our collected data with moderation labels to train the classifier $f$.
%With the collected Reddit comments and their labels, we can train a model to discriminate between removed and non-removed comments.
We use pre-trained transformer based language models as text encoders and add a shallow classifier on top: $y=f_{m}(f_{enc}(x))$, where $f_{enc}$ is the encoder and $f_m$ is the moderation classifier.
To deal with multilingual data, we need to either use a multilingual encoder, that can work with inputs in any language (MLLM), or use a monolingual encoder with machine translation models\footnote{Following \citet{artetxe-etal-2020-translation}, we use two settings with machine translation: 1) \emph{Translate-test}: a classifier is trained on English data, and each non-English language needs to be translated into English during test, and 2) \emph{Translate-train}: translate the English data into each target non-English language, and then do both training and test in that target language. For translation we used models from \citet{TiedemannThottingal:EAMT2020}}.
We only transfer models trained on English data to non-English data and not the other way since 
the number of data points in English are much larger and it could be beneficial to leverage
it for other languages with less data available.

We summarize our experimental results\footnote{Averaged over 3 runs with different random seeds.} in \autoref{tab:classification-original} and observe that the RoBERTa ($67.38$) and XLM-RoBERTa ($66.83$) models outperform offensive language detectors (second model in \autoref{tab:offensive_metric} achieved $61.11$ on AUC) on the English data. This is expected because the moderation training data contains removed comments that are non-offensive, and thus the trained classifiers are able to capture these and make better predictions. 
% In \autoref{sec:examples} we will show examples on this.
We also observe that translate-train performs lower than translate-test in most cases,
%Two possible causes are: 1) The latter has much more training data than the former, which gives better generalization, and 
possibly due to the fact that noise is often introduced during translation and the negative impact is more severe in the training stage than in the test stage. 
% ksikka- Following comment is not reqd
% Thus, it is preferable to keep the training data in its original language and form instead of processing them in ways that might cause lost of information.
We note that although the best performance on most of the test splits is achieved by translate-test, MLLM performance is close to the best in most cases. Considering the challenges of maintaining an extra translation model for every language in real-world applications, MLLM might be a better solution towards building multilingual auto moderators.

We would like to note that the (macro) F1 scores of these models are close to $50$ since the classifier/predictions are overwhelmed by data imbalance (less than $2\%$ of comments are labeled to be removed).
We thus additionally report AUC scores since, being a ranking based metric, it is not affected by data imbalance.
% We would like to note that the (macro) F1 scores of these models are close to $50\%$, 
% because the classifiers are overwhelmed by negative classes that few to none of the predictions are positive, despite that we have used higher weights for the positive training examples. The AUC scores were not affected much since it is a metric of ranking.
For comparison, we also propose additional data splits\footnote{Statistics of the balanced splits can be found in \autoref{tab:statistics_balanced}.} with balanced classes and repeat the same experiments.
As shown in \autoref{tab:classification-balanced}, the performance trends for the balanced set are similar to the original splits, and the F1 scores are clearly above $50$, ranging from $55.10$ to $64.46$ with one exception of $51.99$, which is likely caused by low translation quality in French.

\begin{table*}[t]
    \centering
    \small
    \begin{tabular}{@{}cp{0.7\linewidth}cc@{}}
        \hline
        {\bf Group} & {\bf Examples} & {\bf GT}/$\bm{f_o}$/$\bm{f_m}$ & {\bf freq.} \\ \hline
        \multirow{2}{*}{1)} & \textit{"every f****** one of those dogs ought to be put down."}, \textit{"my god this sub is a perpetual b***h fest. clearly time to unsub"} & \multirow{2}{*}{Y/Y/Y} & \multirow{2}{*}{$7.35\%$} \\
        \hdashline[1pt/2pt]
        \multirow{3}{*}{2)} & \textit{"prove it by sending me \$5 000.00.  that would be nothing for you... so go ahead and do it ..."}, \textit{"how do you explain exploding infections and new variants coming out of south africa where 90\% of the population is vaccinated?"} & \multirow{3}{*}{Y/N/Y} & \multirow{3}{*}{$16.95\%$} \\ 
        \hdashline[1pt/2pt]
        3) & \textit{"f**k \{\{name\}\} and his shorts."}, \textit{"she has such a punchable face"} & Y/Y/N & \multirow{1}{*}{$1.85\%$} \\
        \hdashline[1pt/2pt]
        \multirow{2}{*}{4)} & \textit{"was....was this written by a cat"}, \textit{"depends on the market. \{\{organization\}\} sold more units than switch last month in the uk."} & \multirow{2}{*}{Y/N/N} & \multirow{2}{*}{$23.85\%$} \\
        \hdashline[1pt/2pt]
        \multirow{1}{*}{5)} & \textit{"oh my god you're pathetic lmao"}, \textit{"well you are a moron so i guess it could be worse"} & \multirow{1}{*}{N/Y/Y} & \multirow{1}{*}{$2.15\%$} \\
         \hline
    \end{tabular}
    \caption{Different groups of comments from the English test data. `GT', $f_o$, and $f_m$ are ground truth label, offensiveness score and moderation score, respectively. ``Y'' stands for offensive/removed and ``N'' stands for non-offensive/non-removed. "freq." denote the proportion of each group in the whole subset (2000 samples).}
    \label{tab:test_examples}
\end{table*}

%%%%%%%%%%%%%%%%%%%%%%%%%%%%%%%%%%%%%%%%%%%%%%%%%%%%
%% Experiment - qualitative examples
%%%%%%%%%%%%%%%%%%%%%%%%%%%%%%%%%%%%%%%%%%%%%%%%%%%%
\subsection{Analysis on Error Types}
\label{sec:examples}

We performed additional analysis to understand the errors made by the learned classifier by randomly sampling $1,000$ examples from both removed and non-removed comments and computing their offensiveness\footnote{Average offensiveness score from classifiers in \autoref{existing}} ($f_o$) and moderation scores ($f_m$). We then organize all comments into groups based on their ground truth label and classifier predictions. \autoref{tab:test_examples} shows five of those groups and their frequencies:
1)There are $7.35\%$ examples which are offensive and both $f_o$ and $f_m$ were able to identify them.  
2) There are $16.95\%$ examples on which $f_o$ failed due to their low offensiveness while $f_m$ was able to identify them successfully.
%3) moderation classifier can misclassify examples with subtle offensiveness or when offensive word(s) appears at the end of sentence (f).
3) $f_m$ missed a small portion of examples ($1.85\%$) that are offensive and should be removed (\emph{model-error}).
4) There is a significant portion of data points ($23.85\%$) that were removed by moderators while both classifiers predicted them to be fine (\emph{model-error}). This group represents the majority of removed comments that have violated rules other than the use offensive language.
5) Both classifiers do not agree with the ground truth label on $2.15\%$ of the comments, most of them are highly offensive and should be removed but were actually approved by moderators. These instances indicate noisy labels in our dataset that are possibly caused by moderators' biases (\emph{human-error}).
Based on these observations we conclude that offensiveness classifiers and moderation classifiers have different types of errors and we believe that there is scope for improvement by using models that: 1) can incorporate knowledge from both OLI and moderation tasks and 2) are robust to label noise.

% ksikka: I would not say the following
% Since attacking group of people based on identity violates one of the site-wide rules on Reddit, we attribute this to the variations in moderators' personal preference and biases during the moderation process.

%%%%%%%%%%%%%%%%%%%%%%%%%%%%%%%%%%%%%%%%%%%%%%%%%%%%
%% Experiment - manual analysis of rule violation
%%%%%%%%%%%%%%%%%%%%%%%%%%%%%%%%%%%%%%%%%%%%%%%%%%%%

\begin{table}[t]
    \centering
    \small
    \begin{tabular}{@{}p{0.4\linewidth}lrr@{}}
        \hline
        \textbf{Subreddit Group} & \textbf{Model} & \textbf{AUC} & \textbf{F1} \\
        \hline
        \multirow{2}{\linewidth}{\tt r/naruto} & M-BERT & 51.59 & 47.43\\
        %\cline{2-4}
        & XLM-Ro & 64.48 & 37.84\\
        \hdashline[1pt/2pt]
        \multirow{2}{\linewidth}{{\tt r/naruto} + {\tt r/anime}} & M-BERT & 55.78 & 54.13\\
        %\cline{2-4}
        & XLM-Ro & 31.08 & 46.39\\
        %\hline
        \hline
        \multirow{2}{\linewidth}{\tt r/judaism} & M-BERT &  62.08 & 60.65\\
        %\cline{2-4}
         & XLM-Ro &   50.23 & 33.77\\
        \hdashline[1pt/2pt]
        \multirow{2}{\linewidth}{{\tt r/judaism} + {\tt r/islam}} & M-BERT &  68.27 & 63.89\\
        %\cline{2-4}
         & XLM-Ro &   62.79 & 33.91\\
        \hdashline[1pt/2pt]
        \multirow{2}{\linewidth}{{\tt r/judaism} + {\tt r/islam} + {\tt r/christianity}} & M-BERT & 67.89 & 62.33 \\ 
        %\cline{2-4}
         & XLM-Ro & 49.27 & 34.50\\
        %\hline
        \hline
        \multirow{2}{\linewidth}{\tt r/feminism} & M-BERT & 71.82 & 64.05\\
        %\cline{2-4}
        & XLM-Ro & 51.27 & 38.22\\
        \hdashline[1pt/2pt]
        \multirow{2}{\linewidth}{{\tt r/feminism} + {\tt r/lgbt}} & M-BERT &  73.25 & 67.26\\
        %\cline{2-4}
        & XLM-Ro & 49.67 & 34.10\\
        \hdashline[1pt/2pt]
        \multirow{2}{\linewidth}{{\tt r/feminism} + {\tt r/lgbt} + {\tt r/racism}} & M-BERT & 72.95 & 66.16 \\ 
        %\cline{2-4}
         & XLM-Ro & 60.78 & 33.80\\ 
        \hline
    \end{tabular}
    \caption{Performance of classifiers within groups of subreddits. M-BERT refers to bert-base-multilingual-cased and XLM-Ro refers to XLM-RoBERTa-base.}
    \label{tab:grouping}
\end{table}

\subsection{Manual Analysis of Rule Violations}
\label{sec:manual_rule_vio}
We also manually labeled $145$ removed comments from the {
\tt r/stock} subreddit with the help of $3$ annotators. The annotators were instructed\footnote{
The guide provided to the annotators is shown in \autoref{tab:annotation_guide}} to select one of the five rules a removed comment violated. 
% reason for removing these posts. These reasons were selected based on the rules of the “r/stock” subreddit. 
We decided to drop class $4$ due to a small number of samples. We also dropped class $5$ as this class corresponded to the “not sure” class. We selected examples where annotators had the most agreement and this resulted in $111$ comments across $3$ classes with $36$, $54$, and $21$ comments for the non-civil, missing context, and spam/self-promotion categories (respectively).  
% data distribution shown in \autoref{tab:rule_vio}.
We performed five-fold cross-validation strategy for our experiments by training a logistic regression classifier on textual features extracted using sentence-transformers (all-mpnet-base-v2)~\cite{reimers-2019-sentence-bert}
% \footnote{\href{https://huggingface.co/sentence-transformers/all-mpnet-base-v2}{https://huggingface.co/sentence-transformers/all-mpnet-base-v2}}. 
We obtained an AUC of $91.8\pm1.8$, and the F1 scores for individual classes are $80.56$ (Non-civil), $77.47$ (Missing context), and $61.53$ (Spam/self-promotion). The class-wise numbers match our intuition that examples from the ``non-civil'' class performs the best as it mostly includes offensive comments.
%This experiment shows that it is possible to learn to predict which rule was violated and thus provide some explanation to the user.
This experiment shows that it is possible to learn to predict which rule was violated and thus provide some explanation to the user.

\subsection{Experiments with Partitioning Data}
As observed earlier, the AUC and F1-macro scores for moderation can be relatively low due to the task's complexity. We hypothesize that grouping similar subreddits together may improve the results within the groups.
%Table \ref{tab:grouping} shows the results of grouping subreddits together.
To verify this, we manually selected three groups of subreddits with similar topics and sample both training and test sets based on the grouping.
%For this set of experiments, we limit our training and test sets to instances from the specific subreddit group we target.
%We limit the size of training data to 10K randomly selected samples from within the subreddit group as we empirically verified (omitted for space) that the performance plateaus off after 10K instances.
From Table \ref{tab:grouping}, we observe that grouping subreddits together may result in improved performance.
For example, grouping {\tt r/lgbt} and {\tt r/feminism} together achieves $73.25$ AUC, an improvement of $1.43$ over {\tt r/feminism} alone, and an improvement of $6\sim7$ compared to Table \ref{tab:classification-balanced}.
However, we also observe that grouping more than two subreddits can lead to a drop in performance. For instance, adding {\tt r/racism} to the group of {\tt r/lgbt} and {\tt r/feminism} decreases the AUC slightly to $72.95$.
%This set of experiments suggest that there is a need for an effective data partitioning algorithm so that a task can be split across multiple classsifiers.
This suggest that splitting the moderation task across multiple classifiers could lead to better performance with proper partitioning.

%%%%%%%%%%%%%%%%%%%%%%%%%%%%%%%%%%%%%%%%%%%%%%%%%%%%
%% Experiment - Insights
%%%%%%%%%%%%%%%%%%%%%%%%%%%%%%%%%%%%%%%%%%%%%%%%%%%%
\subsection{Insights on Moderation Dataset and Task}
Based on the qualitative and quantitative evaluations, we summarize our insights:

\noindent{\bf Moderation is a highly imbalanced task.} 
Unlike existing offensive language datasets, which may have $\ge 30\%$ positive examples \cite{mandl2020overview,zampieri-etal-2019-predicting}, the proportion of removed comments in our moderation dataset is $\leq2\%$ 
This is challenging for most classifiers as their prediction will be biased by the majority class (see \autoref{tab:classification-original}).
% training and testing in that a classifier could be overwhelmed by the majority class and the predictions are dominated by that class, 
% or otherwise a classifier could be trained on balanced data and produce many false positive predictions when deployed to real scenarios. 
We believe that this challenge makes our dataset distinct from others because it exposes models to real world scenarios. 

\noindent{\bf Moderation labels are noisy.}
The moderation labels in our dataset are noisy as we found that some highly offensive comments were not removed and some benign comments were removed by the moderators. There could be multiple reasons for this such as human errors, individual biases. We believe this label noise is a characteristic of the task and researchers will be required to design algorithms with this in mind.
% including occasional mistakes made by the moderators and that different moderators might have their own biases or preference of to what extent to adhere to the guidelines.

\noindent{\bf Moderation requires more than offensiveness detection.}
Our experiments reveal that pre-existing datasets with hate speech or offensive language labels are not sufficient for moderation task. This task involves referencing a set of guidelines which include not only being civil, but also other community-specific rules such as no off-topic discussions, no self-promotions, and no low-effort comments. Models trained only on offensive language datasets are able to identify non-civil content but fail to detect other cases of rule violation.

\noindent{\bf Moderation systems need to be adaptive.}
As Reddit communities are self-organized and moderators can make their own rules, many subreddits have very specific rules that may not be common with other subreddits. For example, ``keep it halo'' ({\tt r/halo}), no ``Asking for handouts or transactions'' ({\tt r/personalfinance}), ``Research must be less than 6 months old'' ({\tt r/science}), and ``No medical advice'' ({\tt r/bodyweightfitness}).
A classifier trained on a large dataset like ours could possibly capture general rules and macro-norms, but is unlikely to transfer to novel communities with very specific rules. 
We thus argue that an adaptive and dynamic system that could condition its decisions on given guidelines is needed for successful auto moderation on platforms such as Reddit.

%%%%%%%%%%%%%%%%%%%%%%%%%%%%%%%%%%%%%%%%%%%%%%%%%%%%
%% Experiment - Limitations and opportunities
%%%%%%%%%%%%%%%%%%%%%%%%%%%%%%%%%%%%%%%%%%%%%%%%%%%%

%%%%%%%%%%%%%%%%%%%%%%%%%%%%%%%%%%%%%%%%%%%%%%%%%%%%
%% Conclusions
%%%%%%%%%%%%%%%%%%%%%%%%%%%%%%%%%%%%%%%%%%%%%%%%%%%%
\section{Conclusions}
In this study we discussed the challenges in content moderation, which include but are not limited to:
differences in community rules, subtlety of offensiveness in different contexts or languages, and moderator biases.
Due to these challenges, prior works on offensiveness or hate speech identification are not adequate for solving moderation. % since they tend to ignore comments that are violating rules while being unoffensive.
%To bridge the gap between existing datasets and the goal of content moderation,
Thus, we propose a multilingual dataset consisting of Reddit comments as well as subreddit meta-data including descriptions and rules.
We experimented with baseline transformer based models to verify our assumptions and show that, although moderation is a challenging task, there are also many opportunities for further studies: linking removed comments to violated rules, moderating comments in context, understanding the differences between languages and culture in moderation process, etc. We believe that our work will foster more research in the area of automatic content moderation.
%%%%%%%%%%%%%%%%%%%%%%%%%%%%%%%%%%%%%%%%%%%%%%%%%%%%
%% Ethical considerations
%%%%%%%%%%%%%%%%%%%%%%%%%%%%%%%%%%%%%%%%%%%%%%%%%%%%
\section*{Limitations}
% We would like to point out two limitations of this study: 1) Our experiments using standard supervised learning methods take each comment separately and make predictions independently, while in practice context should also be taken into consideration, and 2) Our model is a universal moderation classifier that did not incorporate community guidelines. A more adaptive model would be able to make predictions conditioned on dynamically changing guidelines among different communities.

% For the above mentioned limitations of our model, we actually have corresponding information provided in the dataset: 1) we have saved metadata of each subreddit including description and rules, and 2) for each comment we also provide its parent post id, which could be used to recover the tree structure of a discussion thread, providing context for a comment. Thus future works could go beyond the baseline approach and incorporate this information to build better models.
\noindent{\bf Moderation within context.} Our experiments using standard supervised learning methods take each comment separately and make predictions independently, while in practice context should also be taken into consideration. When collecting each comment we have recorded its parent post id, which could be used to recover the tree structure of a discussion thread, providing context for a comment.

\noindent{\bf Incorporating community rules.} Our baseline model is a universal moderation classifier that did not incorporate community rules. A more adaptive model would be able to make predictions conditioned on rules that are dynamically changing among different communities. We provide metadata of each subreddit with our dataset that includes a description of the subreddit and subreddit rules so that future research can incorporate this information to build better approaches.

\section*{Ethics Statement}
We have discussed in detail the data selection process while working with different subreddits.
We had approval from our Institutional Review Board (IRB) for collecting social media data.
We only collected publicly available data (comments, submissions, and subreddit metadata) while adhering to Reddit's policy and did not collect any user-related Personal Identifiable Information (PII) (e.g., Date of Birth) intentionally.
To address the possible appearance of PII in public comments, we went through a data cleaning process using scrubadub\footnote{\url{https://scrubadub.readthedocs.io/}} to remove word tokens or phrases that could be a person's name, email, SSN, driver's license number, etc., to further reduce the risk of PII leakage.
We have kept the data and codebase on secure servers that are accessible only by persons involved in this study.
Our dataset will be available to researchers who agree to the terms and conditions of our data-sharing policy approved by the ethical committee of our institution.

\section*{Acknowledgment}
This project was supported by DARPA grant prime OTA No. HR00112290024 (subcontract No. AWD00005100 and SRA00002145). We also acknowledge the Center for Research Computing at the University of Pittsburgh for providing computational resources.

% \section*{Acknowledgements}
% This material is based upon work supported by the Defense Advanced Research Projects Agency (DARPA) under Agreement No. HR00112290024.
% Any opinions, findings and conclusions or recommendations expressed in this material are those of the author(s) and do not necessarily reflect the views of DARPA. 

\bibliography{custom}

\begin{thebibliography}{33}
\expandafter\ifx\csname natexlab\endcsname\relax\def\natexlab#1{#1}\fi

\bibitem[{Artetxe et~al.(2020)Artetxe, Labaka, and
  Agirre}]{artetxe-etal-2020-translation}
Mikel Artetxe, Gorka Labaka, and Eneko Agirre. 2020.
\newblock \href {https://doi.org/10.18653/v1/2020.emnlp-main.618} {Translation
  artifacts in cross-lingual transfer learning}.
\newblock In \emph{Proceedings of the 2020 Conference on Empirical Methods in
  Natural Language Processing (EMNLP)}, pages 7674--7684, Online. Association
  for Computational Linguistics.

\bibitem[{Atwell et~al.(2022)Atwell, Hassan, and Alikhani}]{Atwell2022APPDIAAD}
Katherine Atwell, Sabit Hassan, and Malihe Alikhani. 2022.
\newblock Appdia: A discourse-aware transformer-based style transfer model for
  offensive social media conversations.
\newblock \emph{ArXiv}, abs/2209.08207.

\bibitem[{Basile et~al.(2019)Basile, Bosco, Fersini, Debora, Patti, Pardo,
  Rosso, Sanguinetti et~al.}]{basile2019semeval}
Valerio Basile, Cristina Bosco, Elisabetta Fersini, Nozza Debora, Viviana
  Patti, Francisco Manuel~Rangel Pardo, Paolo Rosso, Manuela Sanguinetti,
  et~al. 2019.
\newblock Semeval-2019 task 5: Multilingual detection of hate speech against
  immigrants and women in twitter.
\newblock In \emph{13th International Workshop on Semantic Evaluation}, pages
  54--63. Association for Computational Linguistics.

\bibitem[{Bhattacharya et~al.(2020)Bhattacharya, Singh, Kumar, Bansal, Bhagat,
  Dawer, Lahiri, and Ojha}]{bhattacharya2020developing}
Shiladitya Bhattacharya, Siddharth Singh, Ritesh Kumar, Akanksha Bansal, Akash
  Bhagat, Yogesh Dawer, Bornini Lahiri, and Atul~Kr Ojha. 2020.
\newblock Developing a multilingual annotated corpus of misogyny and
  aggression.
\newblock \emph{arXiv preprint arXiv:2003.07428}.

\bibitem[{Chandrasekharan et~al.(2018)Chandrasekharan, Samory, Jhaver, Charvat,
  Bruckman, Lampe, Eisenstein, and Gilbert}]{chandrasekharan2018internet}
Eshwar Chandrasekharan, Mattia Samory, Shagun Jhaver, Hunter Charvat, Amy
  Bruckman, Cliff Lampe, Jacob Eisenstein, and Eric Gilbert. 2018.
\newblock The internet's hidden rules: An empirical study of reddit norm
  violations at micro, meso, and macro scales.
\newblock \emph{Proceedings of the ACM on Human-Computer Interaction},
  2(CSCW):1--25.

\bibitem[{Chaudhary et~al.(2021)Chaudhary, Saxena, and
  Meng}]{chaudhary2021countering}
Mudit Chaudhary, Chandni Saxena, and Helen Meng. 2021.
\newblock Countering online hate speech: An nlp perspective.
\newblock \emph{arXiv preprint arXiv:2109.02941}.

\bibitem[{Chung et~al.(2019)Chung, Kuzmenko, Tekiroglu, and
  Guerini}]{chung-etal-2019-conan}
Yi-Ling Chung, Elizaveta Kuzmenko, Serra~Sinem Tekiroglu, and Marco Guerini.
  2019.
\newblock \href {https://doi.org/10.18653/v1/P19-1271} {{CONAN} - {CO}unter
  {NA}rratives through nichesourcing: a multilingual dataset of responses to
  fight online hate speech}.
\newblock In \emph{Proceedings of the 57th Annual Meeting of the Association
  for Computational Linguistics}, pages 2819--2829, Florence, Italy.
  Association for Computational Linguistics.

\bibitem[{Dadvar et~al.(2013)Dadvar, Trieschnigg, Ordelman, and
  Jong}]{dadvar2013improving}
Maral Dadvar, Dolf Trieschnigg, Roeland Ordelman, and Franciska~de Jong. 2013.
\newblock Improving cyberbullying detection with user context.
\newblock In \emph{European Conference on Information Retrieval}, pages
  693--696. Springer.

\bibitem[{Dinakar et~al.(2011)Dinakar, Reichart, and
  Lieberman}]{dinakar2011modeling}
Karthik Dinakar, Roi Reichart, and Henry Lieberman. 2011.
\newblock Modeling the detection of textual cyberbullying.
\newblock In \emph{Proceedings of the International AAAI Conference on Web and
  Social Media}, volume~5, pages 11--17.

\bibitem[{Dosono and Semaan(2019)}]{dosono2019moderation}
Bryan Dosono and Bryan Semaan. 2019.
\newblock Moderation practices as emotional labor in sustaining online
  communities: The case of aapi identity work on reddit.
\newblock In \emph{Proceedings of the 2019 CHI conference on human factors in
  computing systems}, pages 1--13.

\bibitem[{Fiesler et~al.(2018)Fiesler, McCann, Frye, Brubaker
  et~al.}]{fiesler2018reddit}
Casey Fiesler, Joshua McCann, Kyle Frye, Jed~R Brubaker, et~al. 2018.
\newblock Reddit rules! characterizing an ecosystem of governance.
\newblock In \emph{Twelfth International AAAI Conference on Web and Social
  Media}.

\bibitem[{Gautam et~al.(2020)Gautam, Mathur, Gosangi, Mahata, Sawhney, and
  Shah}]{gautam2020metooma}
Akash Gautam, Puneet Mathur, Rakesh Gosangi, Debanjan Mahata, Ramit Sawhney,
  and Rajiv~Ratn Shah. 2020.
\newblock \# metooma: Multi-aspect annotations of tweets related to the metoo
  movement.
\newblock In \emph{Proceedings of the International AAAI Conference on Web and
  Social Media}, volume~14, pages 209--216.

\bibitem[{Geiger and Ribes(2010)}]{geiger2010work}
R~Stuart Geiger and David Ribes. 2010.
\newblock The work of sustaining order in wikipedia: The banning of a vandal.
\newblock In \emph{Proceedings of the 2010 ACM conference on Computer supported
  cooperative work}, pages 117--126.

\bibitem[{Gillespie(2020)}]{gillespie2020content}
Tarleton Gillespie. 2020.
\newblock Content moderation, ai, and the question of scale.
\newblock \emph{Big Data \& Society}, 7(2):2053951720943234.

\bibitem[{Golbeck et~al.(2017)Golbeck, Ashktorab, Banjo, Berlinger, Bhagwan,
  Buntain, Cheakalos, Geller, Gnanasekaran, Gunasekaran
  et~al.}]{golbeck2017large}
Jennifer Golbeck, Zahra Ashktorab, Rashad~O Banjo, Alexandra Berlinger,
  Siddharth Bhagwan, Cody Buntain, Paul Cheakalos, Alicia~A Geller,
  Rajesh~Kumar Gnanasekaran, Raja~Rajan Gunasekaran, et~al. 2017.
\newblock A large labeled corpus for online harassment research.
\newblock In \emph{Proceedings of the 2017 ACM on web science conference},
  pages 229--233.

\bibitem[{Hassan et~al.(2022)Hassan, Atwell, and Alikhani}]{hassan2022studying}
Sabit Hassan, Katherine~J Atwell, and Malihe Alikhani. 2022.
\newblock Studying the effect of moderator biases on the diversity of online
  discussions: A computational cross-linguistic study.
\newblock In \emph{Proceedings of the Annual Meeting of the Cognitive Science
  Society}, volume~44.

\bibitem[{Jhaver et~al.(2017)Jhaver, Vora, and Bruckman}]{jhaver2017designing}
Shagun Jhaver, Pranil Vora, and Amy Bruckman. 2017.
\newblock Designing for civil conversations: Lessons learned from changemyview.
\newblock Technical report, Georgia Institute of Technology.

\bibitem[{Kennedy et~al.(2017)Kennedy, McCollough, Dixon, Bastidas, Ryan, Loo,
  and Sahay}]{kennedy2017technology}
George Kennedy, Andrew McCollough, Edward Dixon, Alexei Bastidas, John Ryan,
  Chris Loo, and Saurav Sahay. 2017.
\newblock Technology solutions to combat online harassment.
\newblock In \emph{Proceedings of the first workshop on abusive language
  online}, pages 73--77.

\bibitem[{Liu et~al.(2019)Liu, Ott, Goyal, Du, Joshi, Chen, Levy, Lewis,
  Zettlemoyer, and Stoyanov}]{liu2019roberta}
Yinhan Liu, Myle Ott, Naman Goyal, Jingfei Du, Mandar Joshi, Danqi Chen, Omer
  Levy, Mike Lewis, Luke Zettlemoyer, and Veselin Stoyanov. 2019.
\newblock Roberta: A robustly optimized bert pretraining approach.
\newblock \emph{arXiv preprint arXiv:1907.11692}.

\bibitem[{Mandl et~al.(2021)Mandl, Modha, Shahi, Jaiswal, Nandini, Patel,
  Majumder, and Sch{\"{a}}fer}]{mandl2020overview}
Thomas Mandl, Sandip Modha, Gautam~Kishore Shahi, Amit~Kumar Jaiswal, Durgesh
  Nandini, Daksh Patel, Prasenjit Majumder, and Johannes Sch{\"{a}}fer. 2021.
\newblock \href {http://arxiv.org/abs/2108.05927} {Overview of the {HASOC}
  track at {FIRE} 2020: Hate speech and offensive content identification in
  indo-european languages}.
\newblock \emph{CoRR}, abs/2108.05927.

\bibitem[{Mubarak et~al.(2022)Mubarak, Hassan, and
  Chowdhury}]{Mubarak2022EmojisAA}
Hamdy Mubarak, Sabit Hassan, and Shammur~A. Chowdhury. 2022.
\newblock Emojis as anchors to detect arabic offensive language and hate
  speech.
\newblock \emph{ArXiv}, abs/2201.06723.

\bibitem[{Pamungkas et~al.(2020)Pamungkas, Basile, and
  Patti}]{pamungkas2020you}
Endang~Wahyu Pamungkas, Valerio Basile, and Viviana Patti. 2020.
\newblock Do you really want to hurt me? predicting abusive swearing in social
  media.
\newblock In \emph{The 12th Language Resources and Evaluation Conference},
  pages 6237--6246. European Language Resources Association.

\bibitem[{Qian et~al.(2019)Qian, Bethke, Liu, Belding, and
  Wang}]{qian2019benchmark}
Jing Qian, Anna Bethke, Yinyin Liu, Elizabeth Belding, and William~Yang Wang.
  2019.
\newblock A benchmark dataset for learning to intervene in online hate speech.
\newblock \emph{arXiv preprint arXiv:1909.04251}.

\bibitem[{Ranasinghe and Zampieri(2021)}]{ranasinghe2021multilingual}
Tharindu Ranasinghe and Marcos Zampieri. 2021.
\newblock Multilingual offensive language identification for low-resource
  languages.
\newblock \emph{Transactions on Asian and Low-Resource Language Information
  Processing}, 21(1):1--13.

\bibitem[{Reimers and Gurevych(2019)}]{reimers-2019-sentence-bert}
Nils Reimers and Iryna Gurevych. 2019.
\newblock \href {https://arxiv.org/abs/1908.10084} {Sentence-bert: Sentence
  embeddings using siamese bert-networks}.
\newblock In \emph{Proceedings of the 2019 Conference on Empirical Methods in
  Natural Language Processing}. Association for Computational Linguistics.

\bibitem[{Roberts(2014)}]{roberts2014behind}
Sarah~T Roberts. 2014.
\newblock \emph{Behind the screen: The hidden digital labor of commercial
  content moderation}.
\newblock University of Illinois at Urbana-Champaign.

\bibitem[{Samory(2021)}]{samory2021positive}
Mattia Samory. 2021.
\newblock On positive moderation decisions.
\newblock In \emph{ICWSM}, pages 585--596.

\bibitem[{Tiedemann and Thottingal(2020)}]{TiedemannThottingal:EAMT2020}
J{\"o}rg Tiedemann and Santhosh Thottingal. 2020.
\newblock {OPUS-MT} — {B}uilding open translation services for the {W}orld.
\newblock In \emph{Proceedings of the 22nd Annual Conferenec of the European
  Association for Machine Translation (EAMT)}, Lisbon, Portugal.

\bibitem[{Vidgen et~al.(2021)Vidgen, Nguyen, Margetts, Rossini, and
  Tromble}]{vidgen-etal-2021-introducing}
Bertie Vidgen, Dong Nguyen, Helen Margetts, Patricia Rossini, and Rebekah
  Tromble. 2021.
\newblock \href {https://doi.org/10.18653/v1/2021.naacl-main.182} {Introducing
  {CAD}: the contextual abuse dataset}.
\newblock In \emph{Proceedings of the 2021 Conference of the North American
  Chapter of the Association for Computational Linguistics: Human Language
  Technologies}, pages 2289--2303, Online. Association for Computational
  Linguistics.

\bibitem[{Wang et~al.(2022)Wang, Bulat, Fujimoto, and Frey}]{wang2022governing}
Hannah~M Wang, Beril Bulat, Stephen Fujimoto, and Seth Frey. 2022.
\newblock Governing for free: Rule process effects on reddit moderator
  motivations.
\newblock In \emph{International Conference on Human-Computer Interaction},
  pages 97--105. Springer.

\bibitem[{Wohn(2019)}]{wohn2019volunteer}
Donghee~Yvette Wohn. 2019.
\newblock Volunteer moderators in twitch micro communities: How they get
  involved, the roles they play, and the emotional labor they experience.
\newblock In \emph{Proceedings of the 2019 CHI conference on human factors in
  computing systems}, pages 1--13.

\bibitem[{Zampieri et~al.(2019)Zampieri, Malmasi, Nakov, Rosenthal, Farra, and
  Kumar}]{zampieri-etal-2019-predicting}
Marcos Zampieri, Shervin Malmasi, Preslav Nakov, Sara Rosenthal, Noura Farra,
  and Ritesh Kumar. 2019.
\newblock \href {https://doi.org/10.18653/v1/N19-1144} {Predicting the type and
  target of offensive posts in social media}.
\newblock In \emph{Proceedings of the 2019 Conference of the North {A}merican
  Chapter of the Association for Computational Linguistics: Human Language
  Technologies, Volume 1 (Long and Short Papers)}, pages 1415--1420,
  Minneapolis, Minnesota. Association for Computational Linguistics.

\bibitem[{Zampieri et~al.(2020)Zampieri, Nakov, Rosenthal, Atanasova,
  Karadzhov, Mubarak, Derczynski, Pitenis, and
  \c{C}\"{o}ltekin}]{zampieri-etal-2020-semeval}
Marcos Zampieri, Preslav Nakov, Sara Rosenthal, Pepa Atanasova, Georgi
  Karadzhov, Hamdy Mubarak, Leon Derczynski, Zeses Pitenis, and
  \c{C}a\u{g}r{\i} \c{C}\"{o}ltekin. 2020.
\newblock {SemEval-2020 Task 12: Multilingual Offensive Language Identification
  in Social Media (OffensEval 2020)}.
\newblock In \emph{Proceedings of SemEval}.

\end{thebibliography}
\bibliographystyle{acl_natbib}

\newpage
\appendix
\section{Appendix}
\label{sec:appendix}
\subsection{Human Annotators}
For better understanding of the moderation data we have performed two studies that involved human annotators: 1) Manual analysis of offensiveness in \autoref{sec:manual_offensiveness}, and 2) manual analysis of rule violations in \autoref{sec:manual_rule_vio}.
In these two studies have used volunteered annotators including employees within our institution and university graduate students, who are all based in North America. We did not use external annotation platforms and services since these are small-scale studies ($\le200$ data points) designed for better understanding of the moderation data and the annotations will not be released with our large main dataset.

\subsection{Temporal Changes in Community Rules}
We scraped the community rules for each subreddit at temporal intervals of 1-2 week(s) and changes in rules have been captured in our data. We checked our entire collection of scraped rules across $7.5$ months and found that rule edits happened at a rate less than $0.93\%$ among all rules from all subreddits per week. Many of those edits are simple changes (e.g., adding/removing periods, capitalizing words) that does not alter their semantic meaning and thus we consider those as not affecting moderation decisions. There are a few cases when the moderators added new rules which would affect moderation decisions of certain types of content. For example, on {\tt r/antiwork}, the rule of "No politicians, no CEOs" was changed to "No politicians, no employers, no landlords, no cops." in between June 27 and July 04. For those cases, developing content moderation models with changing rules would be a good extension of this work.

\subsection{Completeness of the checking later procedure in data collection}
In our preliminary study we found that comment removal happens most frequently during the first $3$ days after being posted. Thus, we set the interval to $1$ week to achieve a reasonable trade-off between completeness and efficiency. We did an additional study by re-scraping the status of an extra set of $452,000$ comments collected during September. Comparing their old labels (checked $7$ days after posting) with new moderation labels (scraped this week, which is more than $2$ months after posting), $92.89\%$ removed comments has already been covered.
%We believe this small portion of $7.11\%$ missing labels for removed comments should not affect the major observations and conclusions in this paper.
Based on our observation of typical subreddits, we also chose the time window of $7$ days to limit domain shifts in the data due to change in topics, moderators or focus of the subreddit.

% A complet list of subreddits:
\begin{table*}[hp]
    \centering
    %\small
    \begin{tabular}{@{}lp{0.7\linewidth}@{}}
        \hline
        {\bf Language} & {\bf Subreddits} \\ \hline
        English & r/anime, r/antivaxxers, r/antiwork, r/birdswitharms, r/bitcoin, r/boardgames, r/bodyweightfitness, r/breadstapledtotrees, r/christianity, r/collapse, r/coronavirus, r/covid19, r/feminism, r/fifthworldproblems, r/funny, r/futurology, r/gadgets, r/games, r/grandpajoehate, r/halo, r/havewemet, r/immigration, r/islam, r/judaism, r/lego, r/lgbt, r/lifehacks, r/mildyinfuriating, r/minecraft, r/naruto, r/news, r/nfl, r/onetruegod, r/personalfinance, r/publicfreakout, r/racism, r/science, r/scifi, r/showerorange, r/showerthoughts, r/space, r/stocks, r/streetwear, r/talesfromcavesupport, r/theoryofreddit, r/tooafraidtoask, r/worldnews, r/xboxseriesx \\ %\hline
        Spanish & r/argentina, r/argaming \\ %\hline
        German & r/de, r/finanzen, r/ich\_iel \\ %\hline
        French & r/rance, r/quebec, r/moi\_dlvv \\
        \hline
    \end{tabular}
    \caption{The list of all subreddits in our dataset.}
    \label{tab:subreddit_list}
\end{table*}

% \subsection{Data collection pipeline}

\begin{figure*}[hp]
    \centering
    \includegraphics[width=0.9\textwidth]{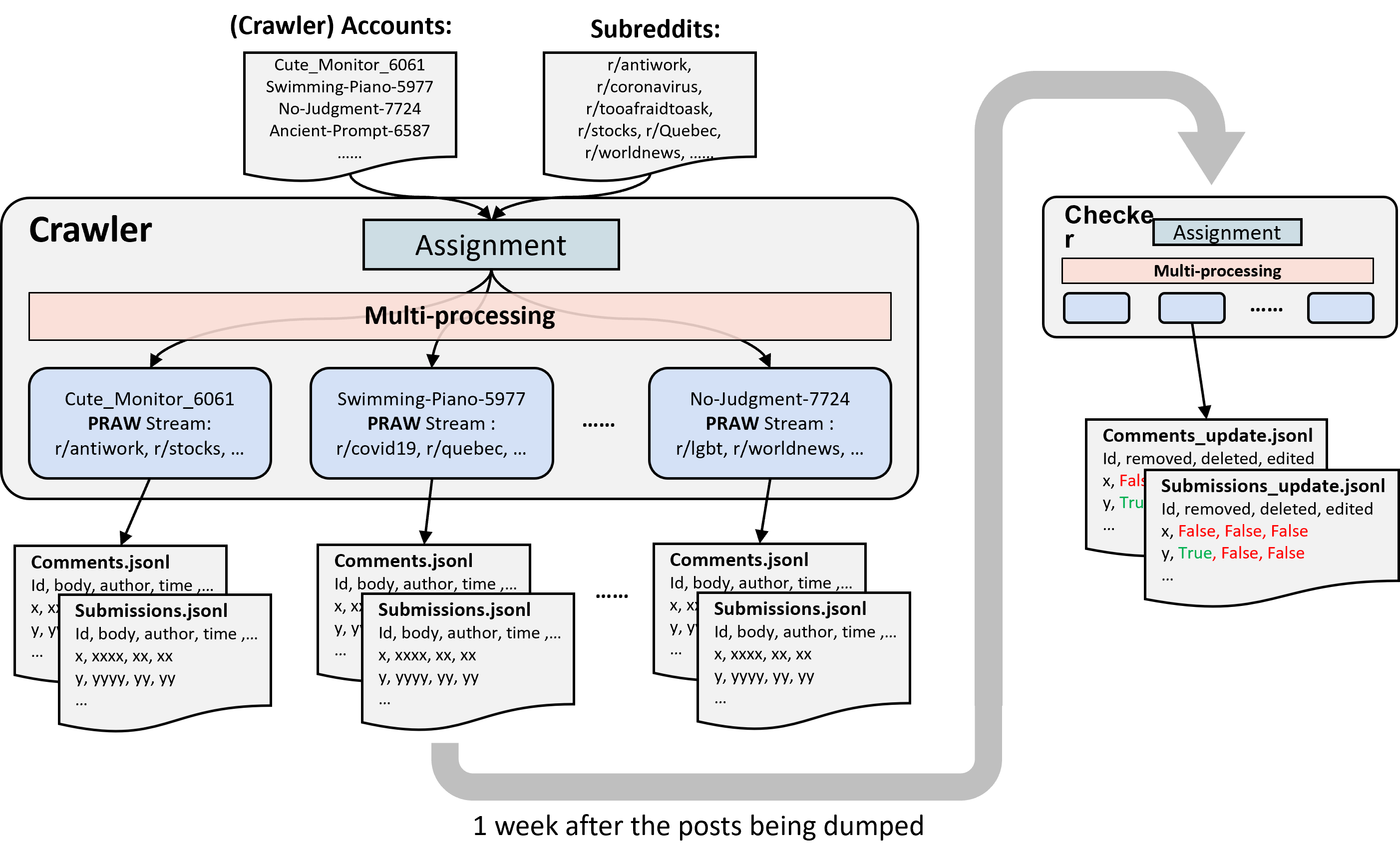}
    \caption{Data collection pipeline: 1) the crawler will assign target subreddits to a number of accounts and streaming comments and submissions concurrently and 2) after a certain period of time (around a week), the checker will go through each record and update if the comment still exists or has been removed/deleted.}
    \label{fig:pipeline}
\end{figure*}

% \subsection{Dataset statistics}
\begin{figure*}[hp]
    \centering
    \includegraphics[width=0.9\linewidth]{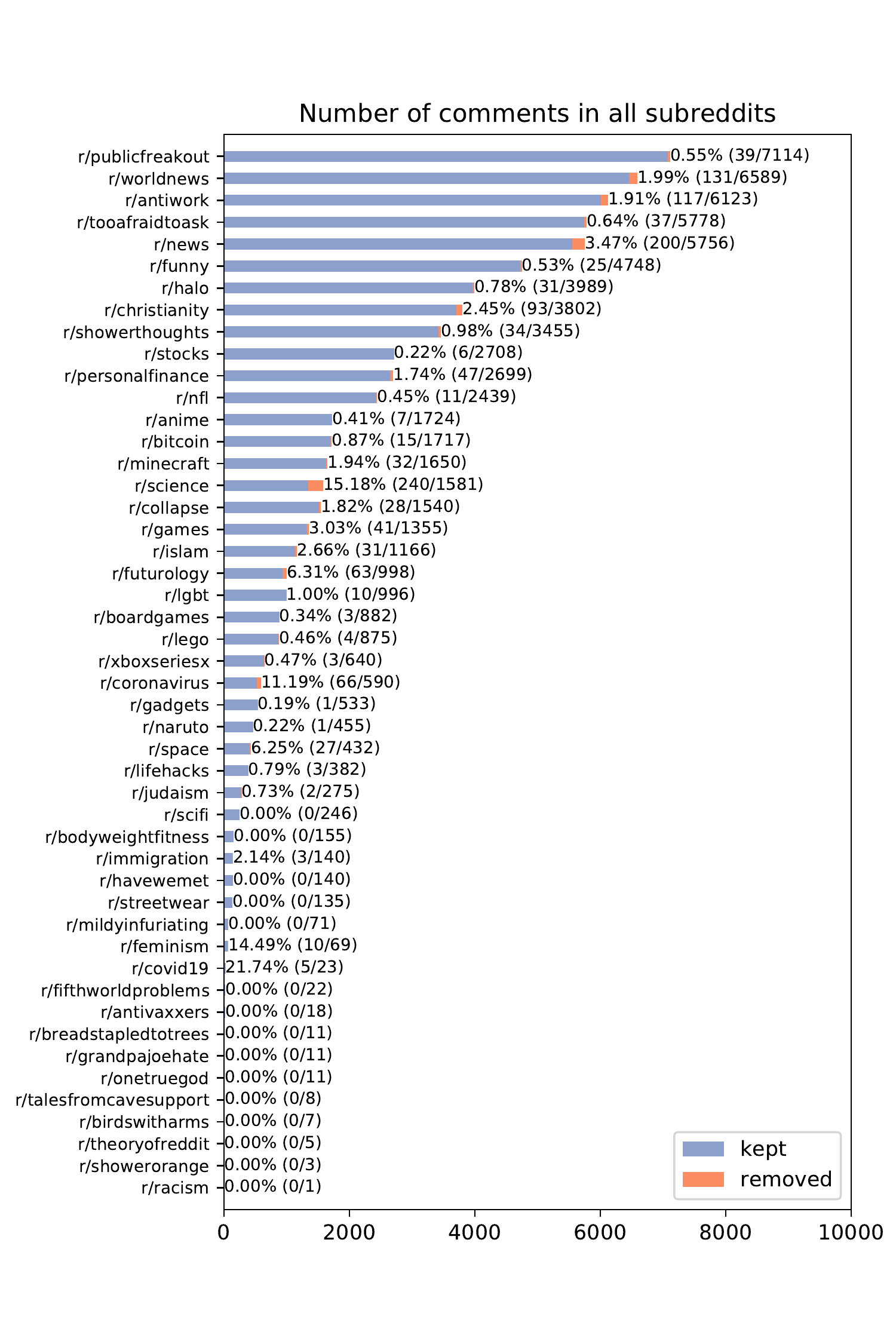}
    \caption{Number of kept and removed comments in the English validation (en-val) data.}
    \label{fig:subreddits-en}
\end{figure*}

\begin{figure*}[hp]
    \centering
    \includegraphics[width=0.9\linewidth]{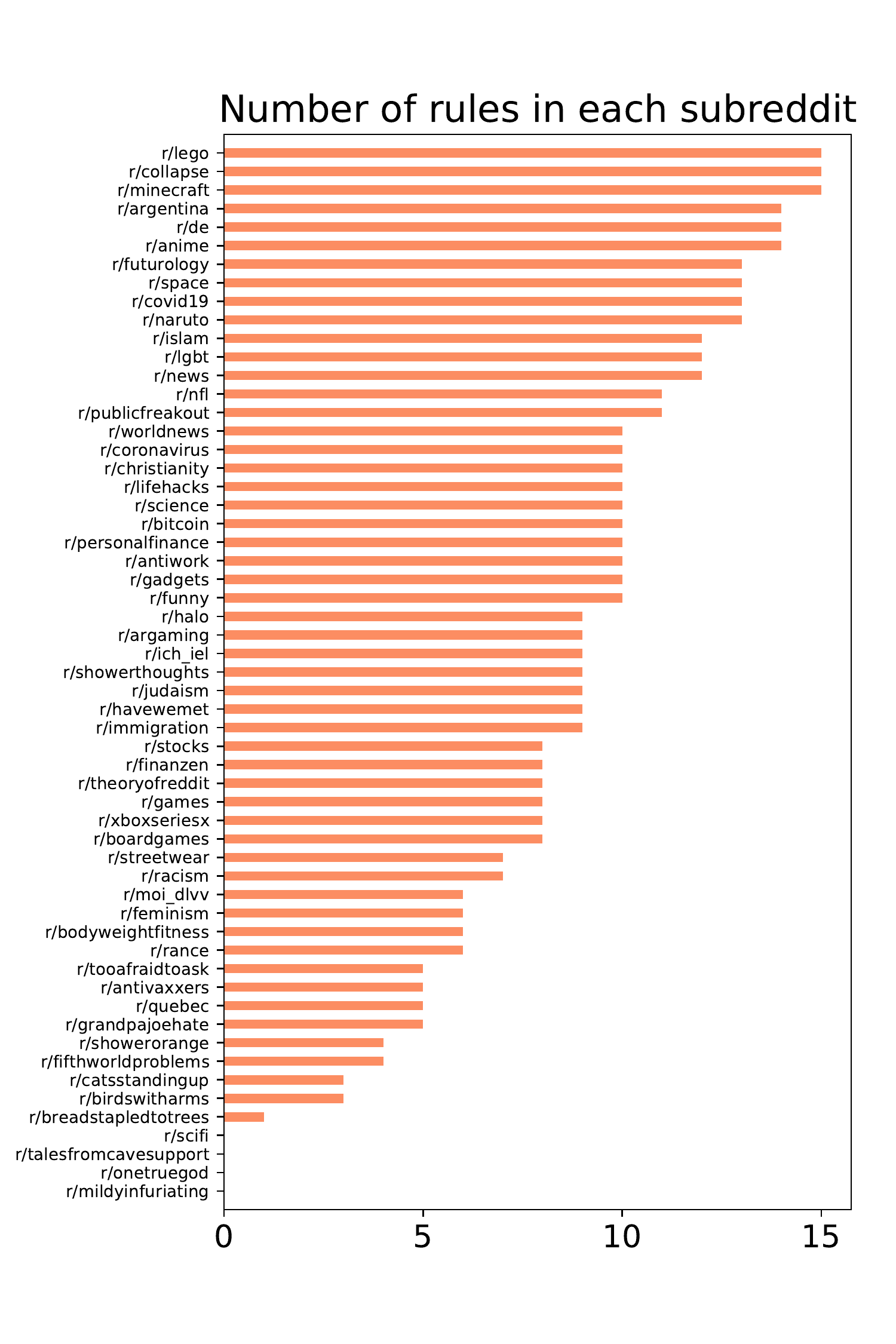}
    \caption{Number of rules in all subreddits.}
    \label{fig:subreddits-rules}
\end{figure*}

\begin{table*}[hp]
    \centering
    %\small
    \begin{tabular}{@{}lp{0.65\linewidth}lr@{}}
        \hline
        & {\bf Comment} & {\bf Subreddit} & $f_o$ \\ \hline
        a & try contacting drewcybersupport on instagram about this he will definitely help you out & {\tt r/minecraft} & $0.01$ \\ %\hline
        b & i think there may still be hope. my company helps clients like yourself get rid of pmi. pm me and we can help you. & {\tt r/personalfinance} & $0.02$ \\ %\hline
        c & i wanna hear ur view on \{\{organization\}\} products and \$sprk tokens. are ya joining anyone? & {\tt r/bitcoin} & $0.02$ \\ %\hline
        d & thanks for the good  constructive conversation  i really did enjoy it. & {\tt r/feminism} & $0.02$ \\ %\hline
        e & maybe it would be easier if you set up a meeting irl. & {\tt r/science} & $0.03$ \\ %\hline
        f & if anyone needs images they are all over /r/transformers right now & {\tt r/lego} & $0.03$ \\ %\hline
        g & so  you have two masters degrees and youre looking to \{\{organization\}\} for help. those degrees are really paying off. & {\tt r/personalfinance} & $0.03$ \\ %\hline
        \hline
        h & no not me it would be another little tiny b***h like you & {\tt r/publicfreakout} & $0.91$ \\ %\hline
        i & ... your piece of s**t car so slow you'll be left behind ...  d**b  silly beta. & {\tt r/tooafraidtoask} & $0.86$ \\ %\hline
        j & we've come so god d**n far in this country ... f**k this s**t!! f**k this country! & {\tt r/antiwork} & $0.82$ \\
        \hline
    \end{tabular}
    \caption{Sampled removed comments and their average offensiveness score from 3 classifiers trained on offensive language identification and hate speech detection datasets.}
    \label{tab:offensiveness_examples}
\end{table*}

\begin{table*}[hp]
%\small
    \centering
    \begin{tabular}{@{}llrrr@{}}
        \hline
        {\bf split} & {\bf language} & {\bf \#sub} & {\bf \#comments} & {\bf removal} \\
        \hline
        \multicolumn{5}{c}{Training data} \\
        %\hline
        en-train & English & $48$ & $1,016,386$ & $47.96\%$ \\
        en-val & English & $48$ & $56,460$ & $47.96\%$  \\
        \hline
        \multicolumn{5}{c}{Test data} \\
        %\hline
        de & German & $3$ & $57,952$ & $50.00\%$ \\
        es & Spanish & $2$ & $18,298$ & $50.00\%$ \\
        en & English & $48$ & $56,466$ & $48.09\%$ \\
        fr & French & $3$ & $5,122$ & $44.57\%$ \\
        \hline
    \end{tabular}
    \caption{Number of subreddits, comments, and percentage of removed comments in the additional balanced data splits.}
    \label{tab:statistics_balanced}
\end{table*}

\begin{table*}[t]
\setlength\tabcolsep{4pt}
    \centering
    \small
    \begin{tabular}{@{}llccccccccc@{}}
         \hline
         \multirow{2}{*}{{\bf Setting}} & \multirow{2}{*}{{\bf Lang.}} & \multicolumn{4}{c}{{\bf AUC}} && \multicolumn{4}{c}{{\bf F1-macro}}\\
         \cline{3-6} \cline{8-11} 
         & & en & de & fr & es && en & de & fr & es \\
         \hline
         MLLM & multi & $66.83_{\pm0.10}$ & $69.42_{\pm0.14}$ & $64.25_{\pm0.14}$ & $63.90_{\pm0.03}$ && $49.60_{\pm0.10}$ & $49.78_{\pm0.00}$ & $49.94_{\pm0.00}$ & $49.83_{\pm0.00}$ \\
         \hdashline[1pt/2pt]
         \multirow{3}{*}{Trans-tr} & de & $60.33_{\pm0.44}$ & $63.34_{\pm0.68}$ & - & - && $49.53_{\pm0.00}$ & $49.78_{\pm0.00}$ & - & - \\
         & fr & $53.47_{\pm0.38}$ & - & $49.96_{\pm0.61}$ & - && $49.53_{\pm0.00}$ & - & $49.94_{\pm0.00}$ & - \\
         & es & $64.16_{\pm0.53}$ & - & - & $63.29_{\pm0.43}$ && $49.58_{\pm0.03}$ & - & - & $49.99_{\pm0.13}$ \\
         \hdashline[1pt/2pt]
         Trans-te & en & $67.38_{\pm0.18}$ & $71.23_{\pm0.02}$ & $71.61_{\pm0.34}$ & $64.33_{\pm0.09}$ && $49.66_{\pm0.05}$ & $50.01_{\pm0.17}$ & $50.36_{\pm0.30}$ & $50.22_{\pm0.30}$ \\
         \hline
    \end{tabular}
    \caption{Experimental results with standard deviation on original splits.}
    \label{tab:classification-original}
\end{table*}

\begin{table*}[t]
\setlength\tabcolsep{4pt}
    \centering
    \small
    \begin{tabular}{@{}llccccccccc@{}}
         \hline
         \multirow{2}{*}{{\bf Setting}} & \multirow{2}{*}{{\bf Lang.}} & \multicolumn{4}{c}{{\bf AUC}} && \multicolumn{4}{c}{{\bf F1-macro}} \\
         \cline{3-6} \cline{8-11}
         & & en & de & fr & es && en & de & fr & es \\
         \hline
         MLLM & multi & $66.14_{\pm0.02}$ & $70.10_{\pm0.07}$ & $68.32_{\pm0.01}$ & $59.81_{\pm0.06}$ && $59.97_{\pm0.51}$ & $64.46_{\pm0.11}$ & $61.15_{\pm0.93}$ & $56.78_{\pm0.06}$ \\
         \hdashline[1pt/2pt]
         \multirow{3}{*}{Trans-tr} & de & $61.10_{\pm0.67}$ & $65.90_{\pm1.22}$ & - & - && $57.24_{\pm1.15}$ & $61.15_{\pm1.17}$ & - & - \\
         & fr & $53.64_{\pm0.37}$ & - & $53.71_{\pm1.40}$ & - && $51.99_{\pm0.98}$ & - & $55.10_{\pm0.56}$ & - \\
         & es & $63.08_{\pm0.14}$ & - & - & $60.52_{\pm0.51}$ && $59.33_{\pm0.26}$ & - & - & $56.87_{\pm0.47}$ \\
         \hdashline[1pt/2pt]
         Trans-te & en & $66.94_{\pm0.05}$ & $70.66_{\pm0.11}$ & $71.57_{\pm0.1}$ & $61.96_{\pm0.03}$ && $61.96_{\pm0.1}$ & $64.16_{\pm0.40}$ & $64.87_{\pm1.30}$ & $58.11_{\pm0.27}$ \\
         \hline
    \end{tabular}
    \caption{Experimental results with standard deviation on balanced splits.}
    \label{tab:classification-balanced}
\end{table*}

\begin{table*}[hp]
    \centering
    %\small
    \begin{tabular}{@{}lc@{}}
    \hline
        {\bf Category} & {\bf number of examples} \\ \hline
        Non-civil & $36$ \\
        Missing context & $54$ \\
        Spam/self-promotion & $21$ \\ \hline
    \end{tabular}
    \caption{Manually annotated rule violations.}
    \label{tab:rule_vio}
\end{table*}

\begin{table*}[hp]
    \centering
    %\small
    \begin{tabular}{@{}cp{0.4\linewidth}cp{0.4\linewidth}@{}}
    \hline
    \multicolumn{4}{l}{\parbox{0.9\linewidth}{\textit{Warning: Due to the nature of this study, this data may contain disturbing content, such as offensive language and hate speech.}}} \\
    \hline
    {\bf Label} & {\bf Rule} && {\bf Examples} \\
    \hline
    1 & not being civil (hate, offensive, insulting, harassment, ...). && (1) you're complete moron. (2) you fuckin loser. (3) have fun staying poor buddy. \\
    \hline
    2 & missing context or effort (Context and effort must be provided; empty posts or empty posts with links will be automatically removed). && (1) https://www.youtube.com/watch?v=... (2) let's GOOOOO. (3) GME. \\
    \hline
    3 & spam or self-promotion (Spam, ads, solicitations (including referral links), and self-promotion posts or comments will be removed). && (1) Checkout my youtube channel. (2) https://www.youbube.com/xxxxxxx. (3) use this link to get \$10 bonus when you register an account at xxxxx. \\
    \hline
    4 & Cryptocoin discussions unrelated to stocks ("I bought bitcoins at coinbase" doesn't count, but "Coinbase sells X amount of bitcoins which is X amount of profit for the company" is fine), penny stocks (including OTC, microcaps, pump \& dumps, low vol pumps and SPACs). && (1) SHIB coin is on fire. (2) \$ATER! get on the train now we are going to the moon! (3) BRQS is the play now, rocketing to \$2 by the end of this week. \\
    \hline
    5 & Not sure why this comment was removed (a comment doesn't seem to have violated any of the rules above, it's hard to say with out context why it was removed) && None \\
    \hline
    \end{tabular}
    \caption{The guide provided to annotators.}
    \label{tab:annotation_guide}
\end{table*}

\end{document}